\pdfoutput=1
\PassOptionsToPackage{table,dvipsnames}{xcolor}
\documentclass[11pt]{article}
\usepackage{wrbench-report}

\usepackage{amsmath,amsfonts,bm}

\def\eqref#1{equation~\ref{#1}}

\def\1{\bm{1}}

\DeclareMathAlphabet{\mathsfit}{\encodingdefault}{\sfdefault}{m}{sl}
\SetMathAlphabet{\mathsfit}{bold}{\encodingdefault}{\sfdefault}{bx}{n}

\usepackage{tikz}
\usetikzlibrary{arrows.meta,positioning,shadows}
\usepackage{wrapfig}
\usepackage{float}

\usepackage{booktabs}
\usepackage{tabularx}
\usepackage{longtable}
\usepackage{enumitem}

\usepackage{pifont}
\usepackage[nosfdefault]{comicneue}

\definecolor{wrblue}{HTML}{5D8DFD}
\definecolor{wrnavy}{HTML}{405D9C}
\definecolor{wrcyan}{HTML}{7DBCC8}
\definecolor{wrsupport}{HTML}{88B06D}
\definecolor{wrmint}{HTML}{AFCFB1}
\colorlet{WRTabdagger}{wrmint!62!white}
\definecolor{wrfail}{HTML}{F05A57}
\definecolor{wrrose}{HTML}{E9A0A3}
\definecolor{wrwarn}{HTML}{E0A64B}
\definecolor{wrgold}{HTML}{F2C36B}
\definecolor{wrviolet}{HTML}{BFA9E6}
\definecolor{wrmuted}{HTML}{777777}
\definecolor{wrbox}{HTML}{F6F7FB}
\definecolor{wrgrid}{HTML}{DCDFE6}
\definecolor{wrplum}{HTML}{B7799E}
\colorlet{wrplumdark}{wrplum!90!black}
\definecolor{wrpink}{HTML}{DBBACB}
\definecolor{wrcream}{HTML}{EDE5E3}
\definecolor{wrsteel}{HTML}{6E87B0}
\definecolor{wrmustard}{HTML}{D6A642}
\colorlet{wrcover}{wrbox}
\colorlet{wrcoveraccent}{wrnavy}
\colorlet{wrcovertext}{black!82}
\colorlet{wrlogotext}{wrnavy}
\definecolor{bench_green}{HTML}{88B06D}
\definecolor{bench_red}{HTML}{F05A57}
\colorlet{bench_orange}{wrmustard}
\definecolor{table_green}{HTML}{E6F0E2}
\definecolor{table_gray}{HTML}{E6E8EE}
\colorlet{WRTableHeaderBg}{wrbox}
\colorlet{WRTableSubheadBg}{wrgrid!42!white}
\colorlet{WRTableFocusBg}{wrplum!10!white}

\usepackage{hyperref}
\hypersetup{
  colorlinks=true,
  linkcolor=wrnavy,
  citecolor=wrnavy,
  urlcolor=wrnavy,
  pdftitle={WRBench: World Models Need More Than Static Scene}
}

\newcommand{\WRTableHead}{\rowcolor{WRTableHeaderBg}}
\newcommand{\WRFocusRow}{\rowcolor{WRTableFocusBg}}
\newcommand{\WRTableNote}[1]{\vspace{2pt}{\scriptsize\emph{Note.} #1\par}}

\newcommand{\focusrow}{\WRFocusRow}
\newcommand{\WRTableSetup}{\small\normalfont\setlength{\tabcolsep}{3pt}\renewcommand{\arraystretch}{1.20}}
\newcommand{\WRTableTightSetup}[1]{\WRTableSetup\setlength{\tabcolsep}{#1}}
\newcommand{\WRWrapTableSetup}[1]{\captionsetup{hypcap=false}\WRTableTightSetup{#1}\renewcommand{\arraystretch}{1.05}}
\newcommand{\WRTableAnalysisSetup}{\WRTableTightSetup{2pt}\renewcommand{\arraystretch}{1.10}}
\newcommand{\WRTableWideSetup}{\footnotesize\normalfont\setlength{\tabcolsep}{1.2pt}\renewcommand{\arraystretch}{1.08}}

\tcbset{findingbox/.style={
    colback=wrbox,
    colframe=wrblue!65!black,
    arc=5pt,
    boxsep=5pt,
    left=8pt,
    right=8pt,
    top=2pt,
    bottom=2pt,
    boxrule=0.7pt,
    drop shadow=gray!35!white,
    enhanced jigsaw
}}
\newcounter{wrfinding}
\newcommand{\finding}[2]{%
  \refstepcounter{wrfinding}%
  \begin{tcolorbox}[findingbox]
  \small\noindent\textbf{Finding~\thewrfinding.\ #1.}\ #2
  \end{tcolorbox}%
}

\DeclareRobustCommand{\benchYes}{\raisebox{-0.05ex}{\textcolor{bench_green}{\large\ding{51}}}}
\DeclareRobustCommand{\benchNo}{\raisebox{-0.05ex}{\textcolor{bench_red}{\large\ding{55}}}}
\DeclareRobustCommand{\benchPart}{
  \tikz[baseline=-0.55ex,scale=0.12]{
    \draw[bench_orange,line width=0.45pt] (0,0) circle (1);
    \begin{scope}
    \clip (0,0) circle (1);
    \fill[bench_orange] (-1,-1) rectangle (0,1);
    \end{scope}
  }
}
\colorlet{GroupTxt}{wrnavy}
\colorlet{GroupBg}{wrblue!10!white}

\colorlet{Besttxt}{wrplum}
\colorlet{Secondtxt}{wrsteel}
\colorlet{Daggertxt}{WRTabdagger}
\colorlet{Sparsetxt}{Daggertxt}

\newcommand{\WRPrimaryGroup}[2]{
  \rowcolor{GroupBg}
  \multicolumn{#1}{@{}l}{\textcolor{GroupTxt}{\textbf{\textit{#2}}}}
}
\newcommand{\WRSubGroup}[2]{
  \rowcolor{WRTableSubheadBg}
  \multicolumn{#1}{@{}l}{\textcolor{GroupTxt}{\textbf{#2}}}
}
\newcommand{\GroupHeader}[1]{\WRPrimaryGroup{10}{#1}}

\newcommand{\WRBest}[1]{\textcolor{Besttxt}{\textbf{#1}}}
\newcommand{\WRSecond}[1]{\textcolor{Secondtxt}{\textbf{#1}}}
\newcommand{\WRDagger}[1]{\textcolor{Daggertxt}{\textbf{#1}}}
\newcommand{\WRSparse}[1]{\textcolor{Sparsetxt}{\ensuremath{\textbf{#1}^{\dagger}}}}
\newcommand{\WRSparseBest}[1]{\textcolor{Besttxt}{\textbf{#1}}\textcolor{Sparsetxt}{\ensuremath{^\dagger}}}
\newcommand{\WRSparseSecond}[1]{\textcolor{Secondtxt}{\textbf{#1}}\textcolor{Sparsetxt}{\ensuremath{^\dagger}}}
\newcommand{\fst}[1]{\WRBest{#1}}
\newcommand{\snd}[1]{\WRSecond{#1}}
\newcommand{\WRGapSig}[1]{\textcolor{wrplumdark}{\textbf{+#1}}}
\newcommand{\WRCondSource}{\mbox{Source-video}}
\newcommand{\WRCondGeom}{\mbox{Geometry-cache}}
\newcommand{\WRCondModel}{\mbox{Model-inferred}}
\newcommand{\WRCondPrompt}{\mbox{Prompt-only}}
\newcommand{\WRHeadOne}[1]{\raisebox{0.26\baselineskip}[0pt][0pt]{#1}}
\newcolumntype{L}[1]{>{\raggedright\arraybackslash}p{#1}}
\newcolumntype{M}[1]{>{\raggedright\arraybackslash}m{#1}}
\newcolumntype{C}[1]{>{\centering\arraybackslash}m{#1}}
\newcolumntype{Y}{>{\centering\arraybackslash}X}

\newcommand{\WRBenchTitleMain}{Current World Models \\ Lack a Persistent State Core}
\newcommand{\WRBenchAuthorList}{%
  Jinpeng~Lu$^{1,2,*}$ \hspace{0.6em}
  Dexu~Zhu$^{2,3,*}$ \hspace{0.6em}
  Haoyuan~Shi$^{1,2}$ \hspace{0.6em}
  Linghan~Cai$^{5}$\\[-0.1em]
  Guo~Tang$^{6}$ \hspace{0.6em}
  Yinda~Chen$^{1,2}$ \hspace{0.6em}
  Jie~Cao$^{3}$ \hspace{0.6em}
  Duyu~Tang$^{4}$\\\hspace{0.6em}
  Yi~Zhang$^{2}$ \hspace{0.6em}
  Yong~Dai$^{2}$ \hspace{0.6em}
  Xiaozhu~Ju$^{2}$%
}
\newcommand{\WRBenchAffiliationList}{%
  $^{1}$ University of Science and Technology of China \quad
  $^{2}$ Beijing Innovation Center of Humanoid Robotics (X-Humanoid) \quad
  $^{3}$ NLPR, Institute of Automation, Chinese Academy of Sciences \quad$^{4}$ Independent Researcher \\[-0.1em]
  $^{5}$ Dresden University of Technology \quad
  $^{6}$ Peking University%
}
\newcommand{\WRBenchContributionList}{$^{*}$ Equal contribution. Work done during the internship at X-Humanoid.}
\newcommand{\WRBenchCodeURL}{https://github.com/JinPLu/WRBench}
\newcommand{\WRBenchProjectURL}{https://jinplu.github.io/WRBench}
\newcommand{\WRBenchCorrespondenceEmails}{%
  \href{mailto:jinplu@mail.ustc.edu.cn}{jinplu@mail.ustc.edu.cn}, %
  \href{mailto:vito.dai@x-humanoid.com}{vito.dai@x-humanoid.com},
  \href{mailto:juxzhu@googlemail.com}{juxzhu@googlemail.com}
  }
\newcommand{\WRBenchDateLabel}{June 2026}
\newcommand{\WRBenchTitleFont}{\wrtitlefont}
\newcommand{\WRBenchDateFont}{\wrdatefont}
\newcommand{\WRBenchMetaLabel}[1]{\wrmetalabel{#1}}
\long\def\WRBenchAbstractText{
World models are increasingly regarded as a decisive step toward artificial general intelligence, yet modeling the physical world demands more than rendering convincing frames on demand: it requires an internal world state that keeps evolving over time, decoupled from observation, so that objects endure and events run to their conclusions whether or not a camera is watching, much as the moon holds to its orbit when no one is looking. This requirement is a blind spot of existing benchmarks, which reward surface properties such as fidelity, motion, and camera controllability while never asking whether a generated world keeps evolving once it is unobserved. We introduce \textbf{WRBench}, the first systematic diagnostic benchmark that treats camera motion as an intervention on observability and resolves evaluation into a human-calibrated chain that asks whether the camera executes the requested interaction, whether the scene stays continuous and identifiable while in view, and whether a returning target remains consistent with the event that was set in motion. Across 9{,}600 videos from 23 models spanning four control paradigms, one finding proves stubborn: current systems maintain the observed world as a tracking shot, resuming a returning target in the state at which it was abandoned rather than advancing the event while it went unseen. Because this failure recurs across control paradigms, model families, and increments of scale, robust world-state evolution does not follow from cleaner imagery, tighter control, richer geometric priors, or sheer parameter count; scaling Wan from $1.3$B to $14$B parameters even lowers re-observed state from $0.66$ to $0.62$, and Lingbot-World, while attaining the strongest visible spatial and state consistency among controllable models at $0.874$ and $0.719$, surrenders control with a requested-camera precision of only $0.468$, the lowest of any controllable model. We therefore argue that the stability of the physical state kernel and the consistency of worldlines under viewpoint intervention should become first-class objectives of world-model design, so that a world model captures how the world will unfold rather than how the next frame appears.}
\newlength{\WRLogoBoxHt}
\newcommand{\WRBenchLogoH}[2]{%
  \begin{minipage}[b]{#1\linewidth}
    \centering
    \vbox to \WRLogoBoxHt{\vfill\hbox to \linewidth{\hfil\includegraphics[height=0.40in]{#2}\hfil}\vfill}%
  \end{minipage}%
}
\newcommand{\WRBenchLogoW}[2]{%
  \begin{minipage}[b]{#1\linewidth}
    \centering
    \vbox to \WRLogoBoxHt{\vfill\hbox to \linewidth{\hfil\includegraphics[width=0.90\linewidth]{#2}\hfil}\vfill}%
  \end{minipage}%
}
\newcommand{\WRBenchLogoRow}{%
  \begingroup
  \WRBenchLogoH{0.13}{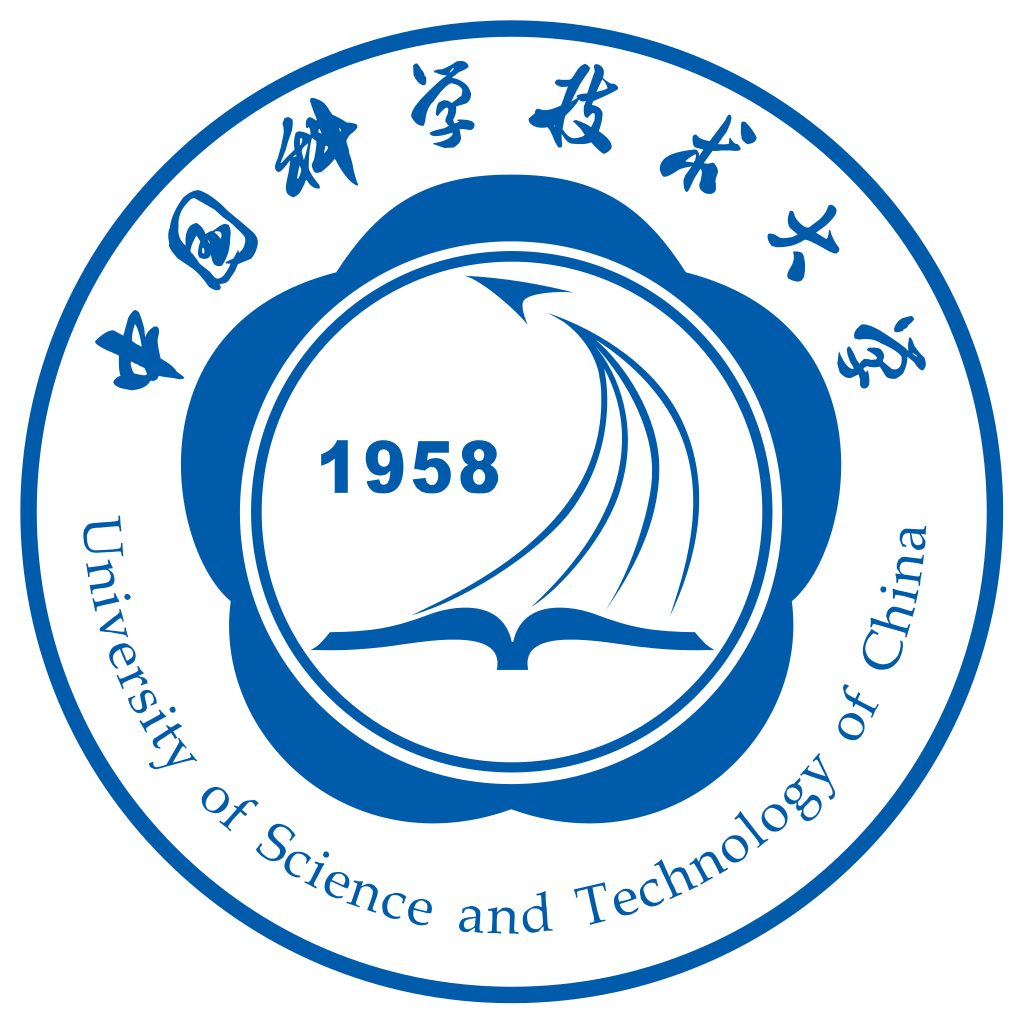}\hfill
  \WRBenchLogoW{0.22}{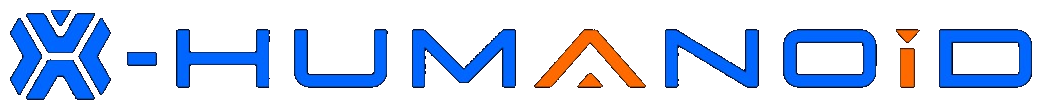}\hfill
  \WRBenchLogoH{0.13}{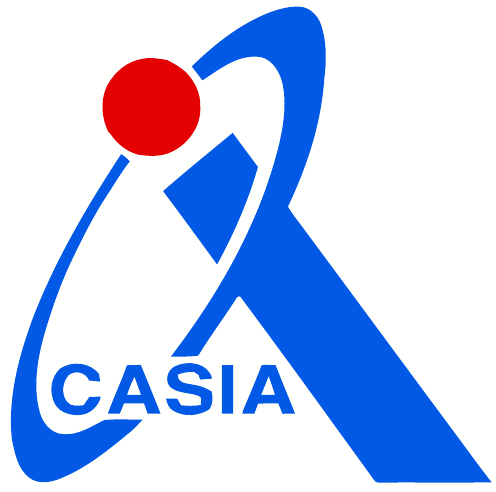}\hfill
  \WRBenchLogoW{0.22}{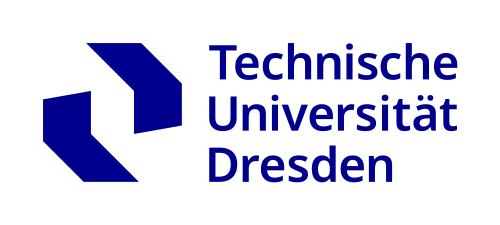}\hfill
  \WRBenchLogoH{0.13}{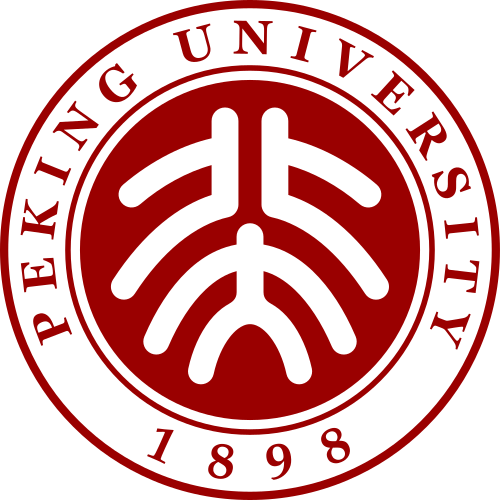}\par
  \endgroup
}
\newcommand{\WRBenchFrontPage}{
  \thispagestyle{empty}
  \vspace*{-0.55in}

  \noindent
  \begin{minipage}[c]{0.72\linewidth}
    \raggedright
    \raisebox{-0.5\height}{\WRBenchLogoRow}
  \end{minipage}%
  \begin{minipage}[c]{0.27\linewidth}
    \raggedleft
    {\WRBenchDateFont\small\color{wrmuted}\WRBenchDateLabel}
  \end{minipage}\par
  \vspace{0.35em}
  \noindent{\color{wrgrid}\rule{\linewidth}{0.8pt}\par}

  \vspace{1.0em}
  \begin{center}
    {{\WRBenchTitleFont\fontsize{21}{25}\selectfont\color{black}%
      \WRBenchTitleMain\par}}
    \vspace{1.25em}

    {\normalsize\rmfamily\color{black}\WRBenchAuthorList\par}
    \vspace{0.22cm}

    {\footnotesize\rmfamily\color{wrmuted}\WRBenchAffiliationList\par}
  \end{center}

  \vspace{0.45em}
  \begin{wrtitlebox}
    \setlength{\parindent}{0pt}
    \setlength{\parskip}{0pt}
    \raggedright
    \nohyphens
    {\small\itshape\color{wrcovertext}%
      ``I like to think the moon is there even if I am not looking at it.''%
      \normalfont\hfill---\,Albert\ Einstein\par}
    \vspace{0.12cm}
    {\small\color{wrcovertext}\WRBenchAbstractText\par}
    \vspace{0.14cm}
    {\setlength{\parskip}{0.06cm}\small
      {\WRBenchMetaLabel{Project Page}\href{\WRBenchProjectURL}{\WRBenchProjectURL}\par}
      {\WRBenchMetaLabel{Code}\href{\WRBenchCodeURL}{\WRBenchCodeURL}\par}
      {\WRBenchMetaLabel{Correspondence}\WRBenchCorrespondenceEmails\par}
    }
    \vspace{0.08cm}
    {\footnotesize\rmfamily\itshape\color{wrmuted}\WRBenchContributionList\par}
  \end{wrtitlebox}

  \pagestyle{fancy}
}

\title{\WRBenchTitleMain}

\begin{document}

\WRBenchFrontPage

\section{Introduction}
\label{sec:intro}

World models are increasingly seen as a key step toward AGI. Video generation has progressed rapidly, and a growing body of work positions generators as visual world models, from interactive environment generation~\citep{bruce2024genie} and physical-AI platforms~\citep{agarwal2025cosmos} to camera- and 4D-aware generation systems~\citep{wang2024motionctrl,he2024cameractrl,ren2025gen3c,zhao2026spatia,zheng2026versecrafter} and memory-based long-horizon generators~\citep{duan2026liveworld}. Yet claiming to be a world model demands far more than producing realistic frames. At minimum, a generator must maintain a world state that evolves continuously over time, independent of observation, not just render plausible frames when watched. The moon orbits the Earth whether or not anyone is looking at it. This is the blind spot of existing benchmarks. Quality-decomposition benchmarks~\citep{huang2024vbench,huang2025vbench++,zheng2025vbench}, temporal and physical benchmarks~\citep{feng2024tc,bansal2025videophy,bansal2025videophy2,meng2024towards,guo2025t2vphysbench,yuan2024chronomagic,sun2025t2v,wang2025your}, and recent world-modeling benchmarks~\citep{li2026worldmodelbench,duan2025worldscore,xu2026worldmark,ying2026wbench,fang2026iworld,ma2026out,zhang2026mbench,ye2026mind} all evaluate fidelity, motion, and camera controllability, but none asks whether the generated dynamic world keeps evolving when the camera looks away.

\begin{figure}[t]
\centering
\includegraphics[width=\linewidth]{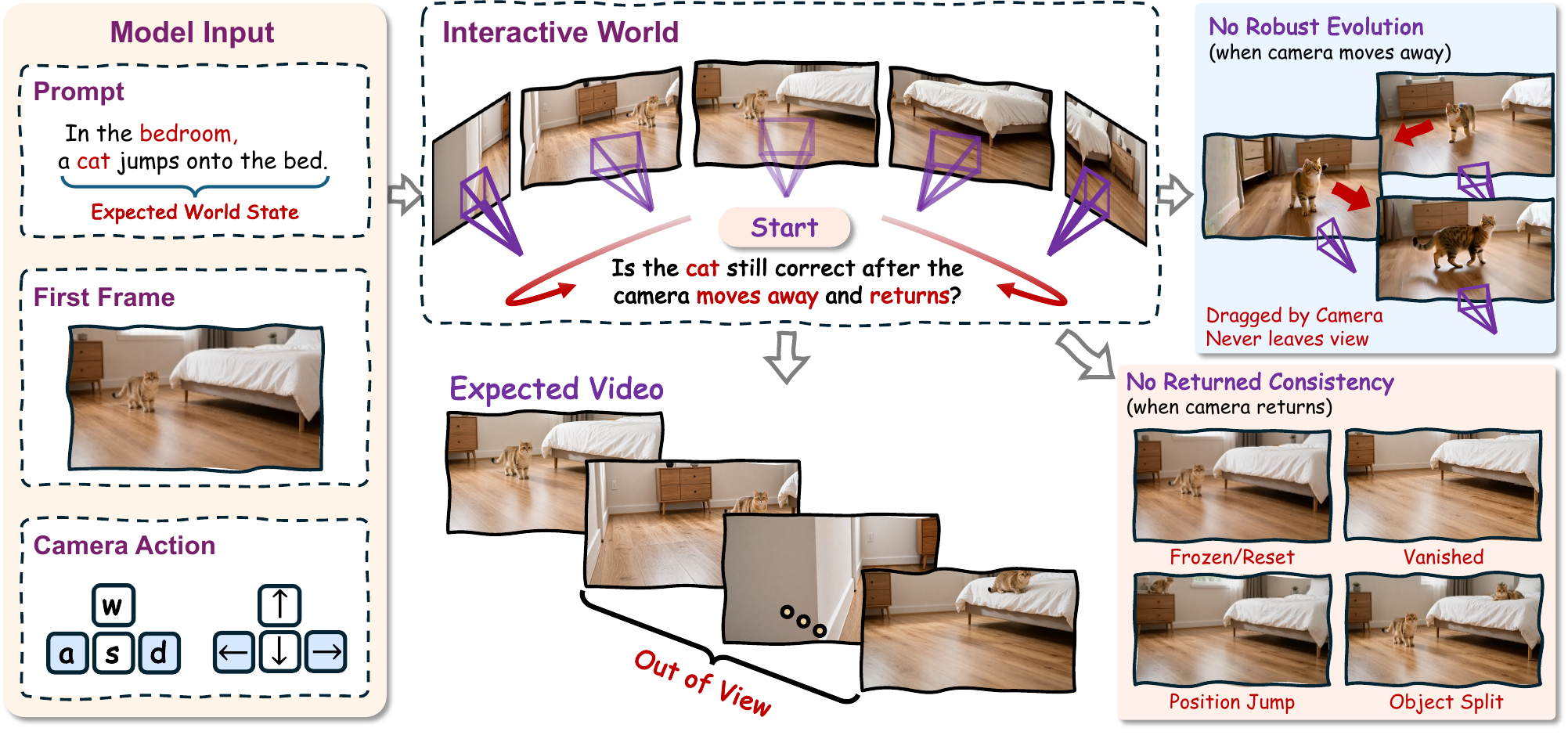}
\caption{\textbf{WRBench Teaser.} A shared scene, event, and viewpoint intervention test whether visible and returned evidence support the same evolving world state. The returned target must preserve the event endpoint, not merely reappear plausibly.}
\label{fig:wrbench_teaser}
\end{figure}

How can this capability be tested? Viewpoint change offers a natural probe: it alters only the angle of observation, not what is happening in the world. As illustrated in Figure~\ref{fig:wrbench_teaser}, the prompt specifies ``in the bedroom, a cat jumps onto the bed,'' with the cat on the floor in the starting frame. The camera then turns away and later returns. If the world state continues to evolve while unobserved, the cat should be on the bed when the camera comes back. In practice, generators exhibit diverse failures (Figure~\ref{fig:wrbench_teaser}, right): the cat may be dragged along with the camera and never leave the view; upon return, it may still be on the floor, may have vanished, may appear in an unexpected position, or may split into duplicates. The difficulty is that these failures stem from entirely different causes: perhaps the camera never turned away, or perhaps it did but the world stopped evolving while unobserved. A single score cannot tell these apart. We call this the \emph{attribution problem} of viewpoint intervention. The capability it tests, that a target's state remains consistent with the event endpoint after leaving and re-entering the view, is what we term \emph{world-state persistence under viewpoint intervention}.

We introduce \textbf{WRBench}, a diagnostic benchmark built around this evidence-attribution problem. WRBench is grounded in \textbf{Natural-25}, a prompt suite of 25 scene families crossed with a four-level event design that factors spatial displacement against state change. Rather than treating camera control as a rendering command, WRBench treats it as an intervention on observability: each test case specifies an initial scene, event, and viewpoint intervention, then asks what claims about the evolving world state are actually supported by the generated video.

Evaluation in WRBench follows a hierarchical diagnostic chain with six dimensions: (i)~requested-camera precision, (ii)~prompt-camera alignment, (iii)~visual integrity, (iv)~visible spatial and state consistency, (v)~re-observation support, and (vi)~re-observed spatial and state consistency. These dimensions progress from control execution to state closure---was the requested trajectory executed, did a prompt-only system follow the intended camera intent, is the frame evidence intact, are spatial relations and action states correct while visible, does the target re-enter the view in judgeable form, and does the returned target preserve the event endpoint. Dimensions (iv) and (vi) each report separate spatial and state scores, but they remain single diagnostic dimensions with shared denominators. All dimensions are reported on their respective denominators; crucially, re-observed-consistency scores in (vi) are computed only on the subset that passes the re-observation gate in (v), so each of the four situations above receives its own diagnostic conclusion rather than being merged into one score.

To enable fair comparison across generators with different architectures and control paradigms, we introduce \textbf{WRBenchLib}, which makes viewpoint-control provenance explicit. It systematically records what form of camera instruction each model received and what condition it actually delivered, supporting condition-level diagnostics across source-video, geometry-cache, model-inferred, and prompt-only controllability paradigms. We also ground the benchmark with 2,547 deduplicated human annotator verdicts that calibrate the automatic evaluation at the same granularity as the benchmark outputs, rather than reducing WRBench to a learned holistic preference score.

We evaluate 23 video generators on 9,600 videos. Analyzing the results along evaluation dimensions, controllability paradigms, and within-family increments, we identify a recurring preservation--access--re-observed-consistency gap: visible quality and re-observation access largely determine whether the return test is posed, while re-observed-state correctness is a separate capability that is not predicted by model scale among the evaluated systems. The failure concentrates on in-place state change rather than relocation. This diagnosis is consistent with the hypothesis that progress requires not more pixels, but a what-memory that records hidden change and a training objective that supervises endpoint persistence.

The contributions are:
\begin{enumerate}
    \item \textbf{Viewpoint-conditioned world-state benchmark.}
    We introduce WRBench to evaluate whether dynamic event state remains attributable under camera-induced changes in observability.
    \item \textbf{Open-source, unified, human-validated toolkit.}
    (a)~We release WRBenchLib, which records delivered viewpoint-control conditions and re-observation judgeability boundaries so that heterogeneous generators are compared by the evidence they actually provide;
    (b)~we release 2,547 deduplicated human annotator verdicts that calibrate and bound the automatic evaluators at the same granularity as the benchmark outputs.
    \item \textbf{Directional diagnosis.} Evaluating 23 models, we show that re-observed-state consistency is a capability distinct from visible fidelity and re-observation access; current scale and architectures do not bind it within the evaluated scope. The diagnosis points toward a what-memory and an endpoint-persistence objective as a direction for future work.
\end{enumerate}

\section{Related Work and Positioning}
\label{sec:related}

\textbf{Decomposed video evaluation.}
WRBench is closest to benchmark suites that turn video generation from a single preference problem into explicit evaluation dimensions. FETV~\citep{liu2023fetv} and EvalCrafter~\citep{liu2024evalcrafter} organize video quality, motion, and text--video alignment into interpretable measurements. VBench establishes the most relevant community-infrastructure pattern: define targeted measurement components, attach tailored evaluators, validate them against human judgments, and release reusable prompts, metrics, generations, and annotations~\citep{huang2024vbench}. VBench++~\citep{huang2025vbench++} and VBench-2.0~\citep{zheng2025vbench} extend this pattern toward image-to-video generation, trustworthiness, controllability, physics, commonsense, and intrinsic faithfulness. WRBench follows the same decomposed, human-calibrated evaluation principle, but changes the measured object: whether world-state evidence remains attributable when viewpoint intervention changes what is visible.

\textbf{World-generation and physical evaluation.}
A second line evaluates generated videos or scenes as candidate worlds. T2V-CompBench~\citep{sun2025t2v}, TC-Bench~\citep{feng2024tc}, ChronoMagic-Bench~\citep{yuan2024chronomagic}, StoryEval~\citep{wang2025your}, VideoPhy~\citep{bansal2025videophy}, VideoPhy-2~\citep{bansal2025videophy2}, PhyGenBench~\citep{meng2024towards}, and T2VPhysBench~\citep{guo2025t2vphysbench} probe prompt composition, event order, metamorphic structure, physical commonsense, and physical-law adherence. WorldModelBench studies world-modeling violations such as instruction following and physics adherence~\citep{li2026worldmodelbench}, while WorldScore is especially relevant because it treats camera trajectory and layout specification as part of the world-generation task~\citep{duan2025worldscore}. WRBench is complementary: it treats camera control not only as an output target or generation condition, but as an evidence intervention. The central question is whether the same event-induced state remains supported as evidence is visible, temporarily unobserved, and returned to a judgeable view.

\textbf{State, memory, and out-of-sight dynamics.}
The closest conceptual neighbors study continuity under interrupted observation, memory, interaction, or revisiting. STEVO-Bench~\citep{ma2026out} tests whether naturally evolving processes continue when observation is interrupted; LiveWorld~\citep{duan2026liveworld} studies out-of-sight dynamics and event permanence; MBench~\citep{zhang2026mbench} and MIND~\citep{ye2026mind} evaluate entity, environment, causal, or action-conditioned memory; and WorldMark~\citep{xu2026worldmark}, WBench~\citep{ying2026wbench}, and iWorld-Bench~\citep{fang2026iworld} study interactive or multi-turn video world models. WRBench shares the motivation that hidden or revisited evidence is a critical world-model test, but it is not a return-only memory benchmark. It separates visible evolution, re-observation support, and re-observed-state consistency. Missing judgeable return evidence is insufficient support for a hidden-state claim; incorrect spatial or event-state evidence after a judgeable return is a re-observed-consistency failure.

\textbf{Heterogeneous viewpoint condition types.}
Recent video generators expose heterogeneous viewpoint interfaces: explicit camera trajectories in MotionCtrl~\citep{wang2024motionctrl} and CameraCtrl~\citep{he2024cameractrl}, novel-view or multi-view generation in ViewCrafter~\citep{yu2024viewcrafter}, CAT4D~\citep{wu2025cat4d}, and GEN3C~\citep{ren2025gen3c}, source-video transformation in ReCamMaster~\citep{bai2025recammaster}, spatial memory in Spatia~\citep{zhao2026spatia}, 4D geometric control in VerseCrafter~\citep{zheng2026versecrafter}, and out-of-sight memory in LiveWorld~\citep{duan2026liveworld}. This heterogeneity is itself an evaluation problem. A model that receives a dense pose trajectory, a model that receives a source video, and a model that receives only a natural-language camera request should not be interpreted as if they were given the same evidence. WRBench therefore records the intended intervention, delivered condition, and generated evidence, so results can be read by viewpoint condition type rather than collapsed into one scalar leaderboard.

\textbf{Table~\ref{tab:benchmark_settings} summarizes this positioning.}
The comparison does not rank the overall value of prior benchmarks; each prior suite covers an important slice of video or world-model evaluation. The table instead records which benchmark targets are explicit enough to enable dynamic world-state evaluation under viewpoint intervention. Here, state robustness means viewpoint/visibility-aware state evidence with separate control, re-observation, and error attribution, and Pathway Diagnostics means metric-, condition-type-, and feature-readout diagnostics.

\begin{table}[t]
\centering
\caption{\textbf{Representative benchmark coverage.} Rows summarize whether prior benchmarks explicitly target dynamic world-state evaluation capabilities.}
\label{tab:benchmark_settings}
\WRTableSetup
\begin{tabularx}{\linewidth}{@{}L{0.26\linewidth}YYYYYY@{}}
\toprule
\WRTableHead
\WRHeadOne{\textbf{Method}} &
\shortstack{\textbf{World}\\\textbf{Dynamics}} &
\shortstack{\textbf{Unified}\\\textbf{Control}} &
\shortstack{\textbf{Visual}\\\textbf{Quality}} &
\shortstack{\textbf{State}\\\textbf{Robustness}} &
\shortstack{\textbf{Evolution}\\\textbf{Consistency}} &
\shortstack{\textbf{Pathway}\\\textbf{Diagnostics}} \\
\midrule
VBench~\citep{huang2024vbench} &
\benchPart & \benchNo & \benchYes & \benchNo & \benchNo & \benchNo \\
VBench-2.0~\citep{zheng2025vbench} &
\benchYes & \benchPart & \benchPart & \benchNo & \benchNo & \benchNo \\
\midrule
WorldModelBench~\citep{li2026worldmodelbench} &
\benchYes & \benchNo & \benchPart & \benchNo & \benchNo & \benchNo \\
WorldScore~\citep{duan2025worldscore} &
\benchYes & \benchPart & \benchYes & \benchNo & \benchNo & \benchNo \\
WorldMark~\citep{xu2026worldmark} &
\benchYes & \benchYes & \benchYes & \benchNo & \benchNo & \benchNo \\
WBench~\citep{ying2026wbench} &
\benchYes & \benchYes & \benchYes & \benchNo & \benchNo & \benchNo \\
iWorld-Bench~\citep{fang2026iworld} &
\benchYes & \benchYes & \benchYes & \benchNo & \benchNo & \benchNo \\
\midrule
MIND~\citep{ye2026mind} &
\benchYes & \benchPart & \benchYes & \benchNo & \benchNo & \benchNo \\
STEVO-Bench~\citep{ma2026out} &
\benchYes & \benchNo & \benchPart & \benchPart & \benchPart & \benchNo \\
LiveBench~\citep{duan2026liveworld} &
\benchYes & \benchPart & \benchPart & \benchPart & \benchPart & \benchNo \\
MBench~\citep{zhang2026mbench} &
\benchYes & \benchPart & \benchNo & \benchPart & \benchPart & \benchNo \\
\midrule
\focusrow \textbf{WRBench (Ours)} &
\benchYes & \benchYes & \benchYes &
\benchYes & \benchYes & \benchYes \\
\bottomrule
\end{tabularx}
\vspace{2pt}
\begin{minipage}{\linewidth}
\raggedright
{\scriptsize\emph{Note.} \textcolor{bench_green}{\ding{51}} indicates full coverage, \benchPart{} partial coverage, and \textcolor{bench_red}{\ding{55}} no coverage; symbols do not rank overall benchmark value.}
\end{minipage}
\end{table}

WRBench is therefore not a survey-style ranking over all video or world-model capabilities. It inherits the human-calibrated evaluation discipline of VBench~\citep{huang2024vbench} and the control-aware world-generation framing of WorldScore~\citep{duan2025worldscore}, but targets a different claim: dynamic world-state attribution under viewpoint intervention. Natural-25, the WRBenchLib generation-provenance layer, judgeability criteria, human calibration, and diagnostic slices across events, cameras, and viewpoint condition types produce bounded evidence for preservation, access, and re-observed-state consistency rather than a single scalar leaderboard.

\section{WRBench Suite}
\label{sec:method}

\begin{figure}[t]
\centering
\includegraphics[width=\linewidth]{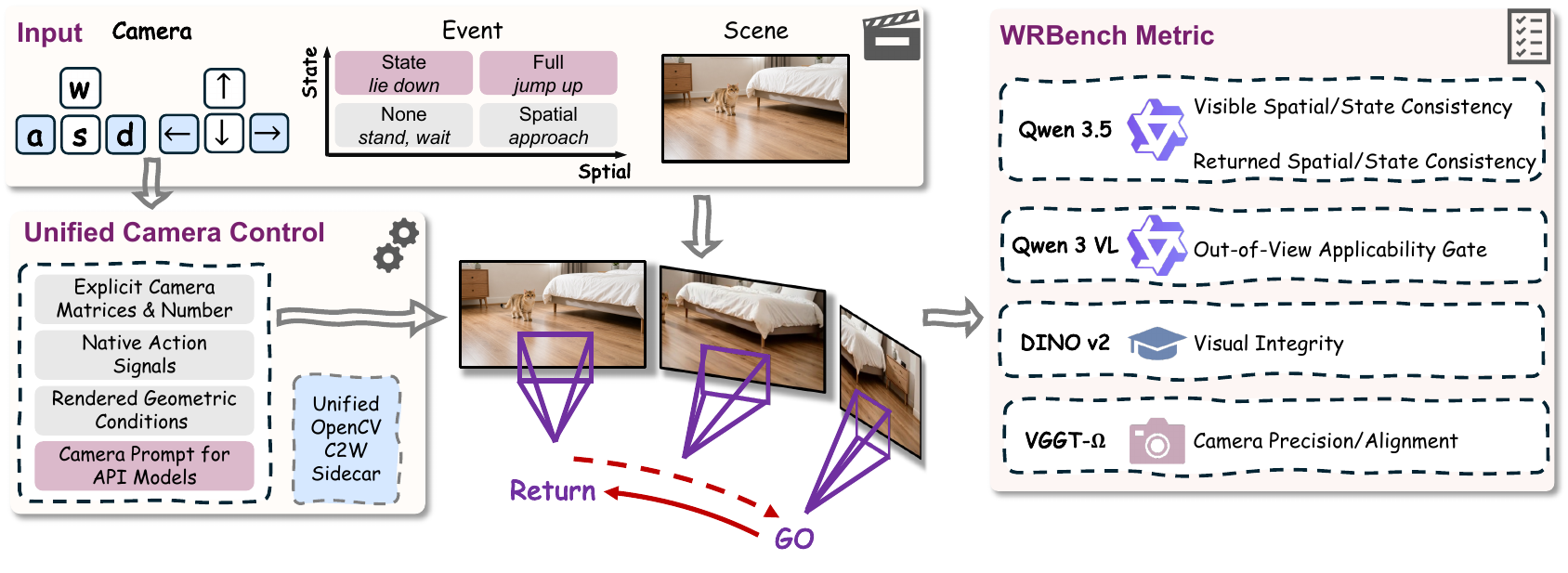}
\caption{\textbf{WRBench method overview.} Natural-25 event-view records supply the scene, event, and viewpoint intervention; WRBenchLib translates each record into a generated video and provenance record per model; the evaluation suite scores six diagnostic dimensions from control execution to re-observed-state consistency; and human preference annotation calibrates each dimension independently.}
\label{fig:wrbench_overview}
\end{figure}

WRBench is a diagnostic benchmark for dynamic world-state evaluation under viewpoint intervention. Following the decomposed-suite design VBench~\citep{huang2024vbench} established for video benchmarks, it is organized in four layers: (1)~a Natural-25 prompt suite that defines the scene and event substrate; (2)~WRBenchLib, a unified toolkit that delivers test conditions to each generator and records what was actually received and produced; (3)~a six-dimensional evaluation suite that scores each generated video from control execution through to re-observed-state consistency; and (4)~human preference annotation that calibrates the automatic evaluators axis by axis. Figure~\ref{fig:wrbench_overview} summarizes the pipeline.

\subsection{Natural-25 Prompt Suite and Event-View Records}
\label{sec:inference_toolbox}
\label{sec:prompt_suite}

This section defines the basic unit of evaluation. WRBench does not score a generated video as a standalone visual-quality sample; it treats the video as evidence for a specific event under a specific viewpoint intervention. Natural-25 supplies the scene and event substrate: 25 scene families crossed with a four-level event design that factors spatial displacement against state change (details in Section~\ref{sec:results}). Each test fixes the same initial observation and event, then varies the viewpoint condition so that the relevant state evidence may remain visible, become temporarily hidden, or return to a judgeable view. Prompts specify the initial scene, event, and viewpoint request without revealing the returned endpoint state, so the generated video itself must carry the evidence for any re-observed-state claim.

The evaluated unit is an \emph{event-view record}, not a prompt or generated clip alone:
\begin{equation}
r_i =
\left(x_i^0, e_i, \tau_i, \nu_i, \pi_i\right).
\label{eq:event_view_record}
\end{equation}
Here $x_i^0$ is the initial observation, $e_i$ the specified event, $\tau_i$ the intended viewpoint intervention, $\nu_i$ the visibility regime, and $\pi_i$ the prompt or interface variant. Spatial evidence covers positions, paths, contacts, supports, and anchors; state evidence covers actions, poses, object states, and events.

\subsection{WRBenchLib: A Unified Toolkit for Scenarios, Events, and Camera Control}
\label{sec:wrbenchlib_toolkit}

With the test unit defined, the next question is how different generators receive it. Some models accept explicit camera trajectories, others take a source video, and others receive only a natural-language prompt. Forcing all models into one interface would make the comparison unfair. WRBenchLib addresses this by translating each event-view record into the form a given generator can consume and recording what was actually delivered, so heterogeneous models are compared on the evidence they received rather than on a single control interface.

WRBenchLib supplies the generation-provenance map for each model $m$:
\begin{equation}
z_{i,m}=\Phi_m(r_i)
=
\left(u_{i,m}, d_{i,m}, v_{i,m}, \eta_{i,m}\right),
\label{eq:wrlib_mapping}
\end{equation}
where $u_{i,m}$ is the model-specific input (trajectory, source video, geometric condition, or prompt), $d_{i,m}$ the condition that was actually delivered, $v_{i,m}$ the generated video, and $\eta_{i,m}$ a provenance record that logs the full pipeline for reproducibility. WRBenchLib records all four components before any evaluation dimension is scored.

\subsection{Evaluation Dimension Suite}
\label{sec:dimension_suite}

With the test unit and delivery layer in place, WRBench scores each generated video along six diagnostic dimensions, matching the chain introduced in Section~\ref{sec:intro}:
\begin{enumerate}[label=(\roman*),leftmargin=*,itemsep=2pt,topsep=4pt]
\item \emph{Requested-camera precision.}
For models that receive an explicit trajectory, did the generated video follow the requested viewpoint path?
\item \emph{Prompt-camera alignment.}
For prompt-only or API models that do not receive a trajectory, did the video follow the intended camera direction (e.g.\ common-yaw, static-hold)?
\item \emph{Visual integrity.}
Is the frame evidence structurally intact and readable, or are there artifacts such as cuts, disappearance, identity drift, or structural collapse that would make downstream judgments unreliable?
\item \emph{Visible spatial and state consistency.}
While the target remains in view, are its spatial relations (position, contact, support) and action/event states correct? Reported as separate spatial and state scores.
\item \emph{Re-observation support.}
After a nontrivial hidden interval, does the target return to the field of view in a form clear enough to be scored? This dimension decides whether the re-observed-state test is posed at all.
\item \emph{Re-observed spatial and state consistency.}
On the subset where (v) is satisfied, does the returned target preserve the expected spatial relation and event endpoint? Again reported as separate spatial and state scores.
\end{enumerate}
Dimensions (i)--(v) are scored over their respective full denominators; dimension (vi) is conditional on (v), so that absence of judgeable return evidence is recorded as insufficient support rather than as a passed or failed state test.

\subsection{Evaluation Method Suite}
\label{sec:evaluation_method_suite}

The previous section defines what WRBench measures; this section describes how each dimension is scored, in the same order.

\paragraph{Camera control (i--ii).}
Dimension (i) targets models that receive explicit trajectories: it compares the requested viewpoint path against the trajectory recovered from the generated video using VGGT-$\Omega$~\citep{wang2026vggt}, built on VGGT~\citep{wang2025vggt}. Dimension (ii) targets prompt-only and API models that do not receive trajectories: it uses CamAlign and static-hold diagnostics to test whether the video follows the intended camera direction (common yaw) rather than full pose recovery. Because the two dimensions address different control interfaces, they are never merged into one control score.

\paragraph{Visual integrity (iii).}
Before scoring any world-state dimension, WRBench checks whether the frame evidence is trustworthy. Visual integrity uses a fixed DINOv2 feature proxy~\citep{oquab2023dinov2}:
\begin{equation}
  I_{\mathrm{vis}}(v) =
  \min\left(s_{\mathrm{global}}(v), s_{\mathrm{local}}(v)\right).
\end{equation}
A low value marks unreliable visual evidence such as cuts, disappearance, identity drift, or local structural collapse. Implementation details are in Appendix~\ref{app:visual-integrity-implementation}.

\paragraph{Visible consistency (iv).}
While the target evidence remains in view, WRBench probes whether spatial relations and action states match the specified event. The probes are prompt-conditioned yes/no questions scored by Qwen-3.5-9B. The context $c$ bundles the scene, event, target object or relation, camera condition, visibility regime, and a dimension-specific rubric. Probe templates and their polarity are fixed before model measurement. For video $v$, context $c$, and question $q$, the evaluator produces a yes probability from yes/no token scores:
\begin{equation}
p_{\mathrm{yes}}(q \mid v,c)
=
\frac{\exp L_{\mathrm{yes}}(q,v,c)}
{\exp L_{\mathrm{yes}}(q,v,c)+\exp L_{\mathrm{no}}(q,v,c)} ,
\end{equation}
where $L_{\mathrm{yes}}$ and $L_{\mathrm{no}}$ are log-sum-exp scores over allowed yes/no answer tokens. Positive probes $\mathcal{P}^{+}_a$ ask whether intended evidence is present; negative probes $\mathcal{P}^{-}_a$ ask whether counterevidence is absent:
\begin{equation}
e_a(q,v,c)=
\begin{cases}
p_{\mathrm{yes}}(q \mid v,c), & q \in \mathcal{P}^{+}_a,\\
1-p_{\mathrm{yes}}(q \mid v,c), & q \in \mathcal{P}^{-}_a .
\end{cases}
\end{equation}
Visible spatial and visible state measurements in dimension (iv) average the polarity-adjusted probes:
\begin{equation}
M_a(v,c)
=
\frac{1}{|\mathcal{P}^{+}_a| + |\mathcal{P}^{-}_a|}
\sum_{q \in \mathcal{P}^{+}_a \cup \mathcal{P}^{-}_a}
e_a(q,v,c).
\end{equation}
Thus dimension (iv) asks whether the event is wrong while evidence remains observable.

\paragraph{Re-observation support and re-observed consistency (v--vi).}
Dimensions (v) and (vi) address evidence that leaves the field of view and later returns --- the distinctive design target of WRBench. Re-observation support (v) records whether the re-observed-state test is even applicable: the target evidence must become temporarily hidden or unjudgeable, later return to the observable field, and be identifiable enough for re-observed scoring. Re-observed spatial and re-observed state consistency in dimension (vi) reuse the same probe form as (iv), but only after judgeable re-observation is established. Let
\begin{equation}
\mathcal{S}_{\mathrm{return}}=\{a_{\mathrm{rsp}},a_{\mathrm{rst}}\}\ \text{and}\ R_a(v,c)=H(v,c)\land U(v,c)\land J_a(v,c),
\label{eq:returned_judgeability}
\end{equation}
where $H$ marks a nontrivial hidden or unjudgeable interval, $U$ marks return to the observable field, and $J_a$ marks enough identifiable returned evidence for metric $a$. These judgeability predicates are evaluated by a separate VLM gate (Qwen-3-VL-Instruct-8B), kept distinct from the Qwen-3.5-9B scoring evaluator so that re-observation applicability is established before any re-observed-consistency score is computed. The re-observed measurements are then
\begin{equation}
M_a(v,c)=
\begin{cases}
\frac{1}{|\mathcal{P}^{+}_a| + |\mathcal{P}^{-}_a|}
\sum_{q \in \mathcal{P}^{+}_a \cup \mathcal{P}^{-}_a}
e_a(q,v,c), & R_a(v,c)=1,\\
\mathrm{NA}, & R_a(v,c)=0,
\end{cases}
\qquad a\in\mathcal{S}_{\mathrm{return}}.
\label{eq:returned_score}
\end{equation}
Dimension (vi) therefore asks whether the endpoint remains coherent after temporary non-observation; dimension (v) decides whether that question can be asked at all.

\paragraph{Aggregation.}
The six dimensions are kept as separate diagnostic outputs rather than collapsed into a single leaderboard score. For model $m$ and metric $a$, let $S_a(v_{i,m},c_i)$ denote whether evidence supports measuring that metric. Camera precision (i), prompt-camera alignment (ii), visual integrity (iii), and visible metrics (iv) use their full applicable denominators, with visual integrity reported separately as an evidence-quality floor. Re-observation support (v) is reported as a support rate over all generated outputs; re-observed metrics in (vi) set $S_a=R_a$ and are averaged only on judgeable rows. WRBench therefore reports both support rates and conditional metric values:
\begin{equation}
\begin{aligned}
\mathcal{D}_{m,a}&=\{i:S_a(v_{i,m},c_i)=1\},\\
\rho_{m,a}&=\frac{|\mathcal{D}_{m,a}|}{|\mathcal{I}_a|},\\
\bar{M}_{m,a}&=
\begin{cases}
\frac{1}{|\mathcal{D}_{m,a}|}
\sum_{i\in\mathcal{D}_{m,a}} M_a(v_{i,m},c_i), & |\mathcal{D}_{m,a}|>0,\\
\mathrm{NA}, & |\mathcal{D}_{m,a}|=0.
\end{cases}
\end{aligned}
\label{eq:support_aggregation}
\end{equation}
The resulting diagnostic profile therefore keeps view failure, unreadable evidence, visible event error, missing re-observation support, and re-observed-state error as distinct, interpretable outcomes rather than averaging them away.

\subsection{Human Preference Annotation}
\label{sec:human_validation}
\label{sec:vlm_human_alignment}

Automatic metrics measure observable evidence but cannot serve as ground truth without human calibration. Rather than collecting a single holistic preference per pair, WRBench asks annotators to judge each of the six diagnostic dimensions independently: requested-view execution, prompt-camera alignment, visual integrity, visible world-state consistency, re-observation support, and re-observed consistency. Each comparison asks which video better satisfies one claim on one axis. For re-observed dimensions, annotators first decide whether hidden-and-returned evidence is judgeable; re-observed-consistency labels are interpreted only on that subset. WRBench reports prevalence-robust AC1~\citep{gwet2008computing} for human agreement and keeps nominal $\kappa$ and ordinal/Krippendorff-style $\alpha$~\citep{krippendorff2011computing} in Appendix~\ref{app:human-alignment-notes}.

For a pair $p=(v^A_p,v^B_p)$ and metric $a$, let $y_{p,a}\in\{-1,0,1\}$ denote the ordered human label. With $\mathcal{C}_a$ denoting pairs for which both human and automatic evidence are defined, WRBench reports rank alignment and a thresholded decision check:
\begin{equation}
\begin{aligned}
\Delta_{p,a} &= M_a(v^A_p,c_p)-M_a(v^B_p,c_p),\\
\rho_a &= \mathrm{Spearman}\left(
\{y_{p,a}\}_{p\in\mathcal{C}_a},
\{\Delta_{p,a}\}_{p\in\mathcal{C}_a}
\right),\\
\hat{y}_{p,a}(\tau_a) &=
\begin{cases}
1, & \Delta_{p,a}>\tau_a,\\
0, & |\Delta_{p,a}|\leq\tau_a,\\
-1, & \Delta_{p,a}<-\tau_a.
\end{cases}
\end{aligned}
\label{eq:human_alignment_bridge}
\end{equation}
The thresholds $\tau_a$ are fixed before reporting and create a tie region around small measurement differences; they are not tuned per model or per result table.

\section{Experiment}
\label{sec:results}

\begin{figure}[t]
\centering
\begin{minipage}[c]{0.48\linewidth}
\centering
\includegraphics[width=\linewidth]{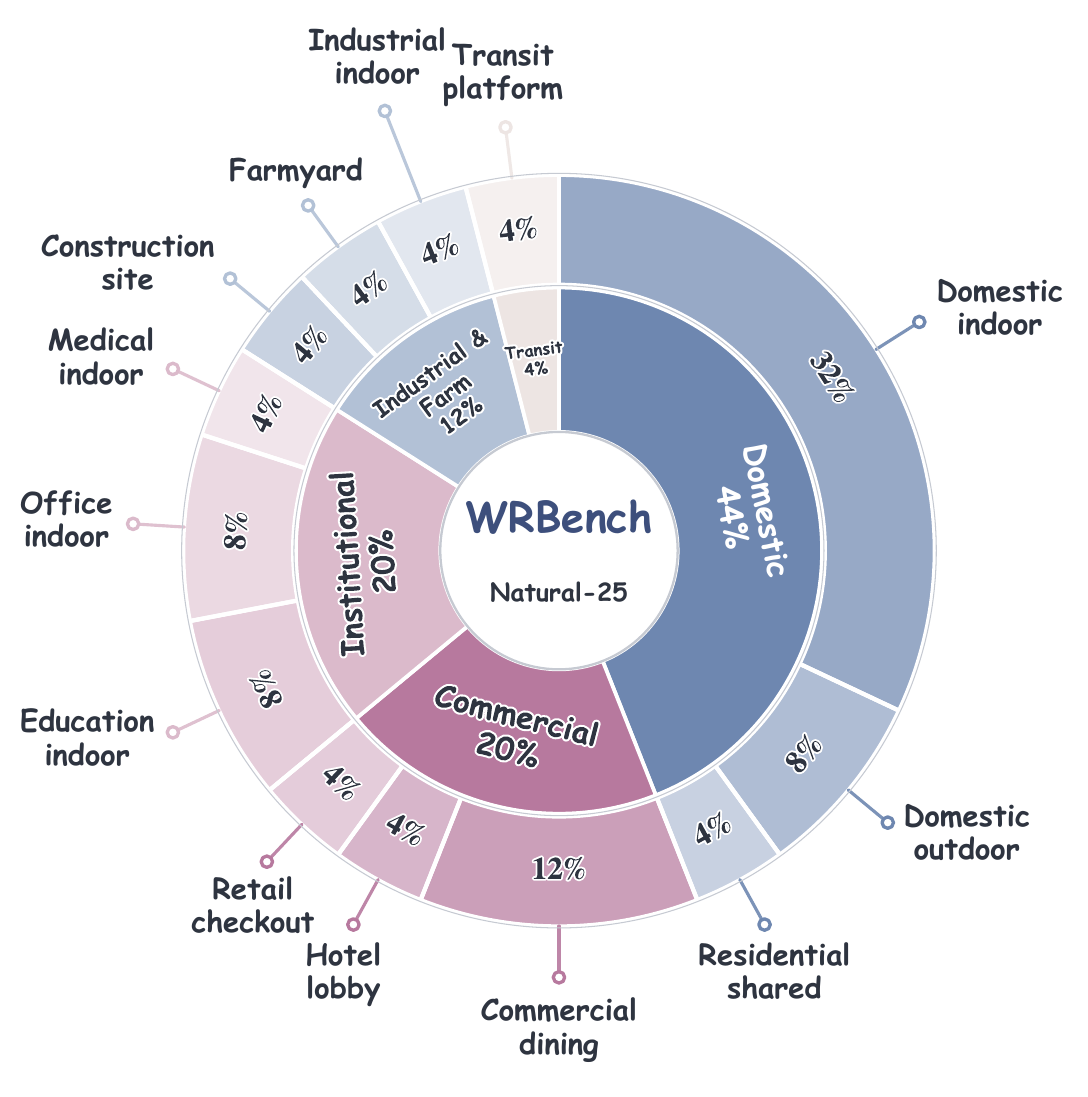}
\end{minipage}\hfill
\begin{minipage}[c]{0.52\linewidth}
\centering
\includegraphics[width=\linewidth]{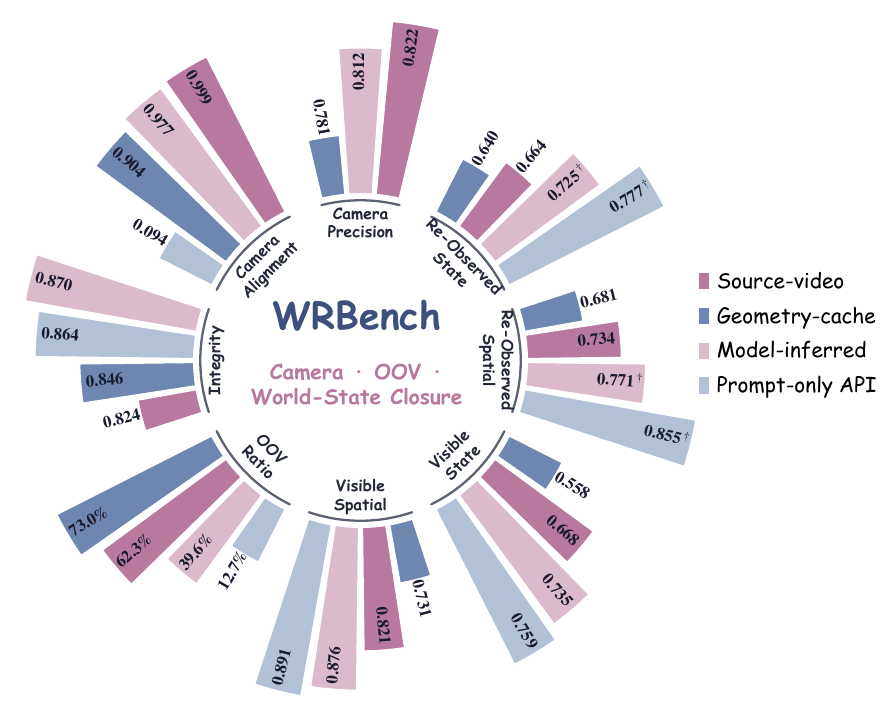}
\end{minipage}
\caption{\textbf{Benchmark coverage and diagnostic frontiers.} Left panel shows Natural-25 scene coverage; right panel shows best observed diagnostic values by viewpoint condition type. Re-observed metrics are conditional on judgeable re-observation, separating access from re-observed correctness.}
\label{fig:wrbench}
\end{figure}

WRBench evaluates 23 models over 9,600 generated videos as a diagnostic profile for dynamic world-state evaluation under viewpoint changes. The profile separates preservation, access, and re-observed consistency. Visual integrity and visible spatial/state consistency are measured over the generated-video evidence where the relevant content is observable. Requested-camera precision is measured on 7,500 certified local requested-control rows, while prompt-camera alignment is a separate common-yaw diagnostic. Re-observed spatial and state consistency are conditionally aggregated over the 2,073 outputs with judgeable hidden-and-returned evidence.

Figure~\ref{fig:wrbench} summarizes the benchmark. The left panel charts Natural-25 coverage as a balanced factorial grid rather than a naturalistic frequency sample. It spans 25 scene families drawn from 19 distinct venues, ranging over indoor and outdoor settings and over human and animal actors, and it crosses every family with a four-level event design that factors spatial change against state change. Domestic indoor scenes form the largest single group, at 8 of the 25 families, yet no family dominates the aggregate, since each contributes an equal share. The suite therefore probes breadth of world state under a controlled design rather than rewarding memorization of frequent scenes. The right panel reports condition-type diagnostic frontiers, the best observed value for each viewpoint condition type on each evaluation dimension. These frontiers preview the main result pattern: current systems preserve visible continuation, instantiate camera motion, expose the hidden target again, and preserve the returned event endpoint at different strengths.

Table~\ref{tab:main-results} is the model-level WRBench diagnostic profile. Its rows are grouped by viewpoint condition type: source-video, geometry-cache, model-inferred, and prompt-only conditions. WRBenchLib records the intended viewpoint intervention, delivered condition, generated video, and measurement evidence before aggregation, so each row is read through the evidence path actually delivered to the benchmark. The central question is endpoint binding: when the camera leaves and returns, does the target-relative displacement, spatial relation, contact outcome, or state change remain bound to the same evolving world state?

\begin{table}[!tbp]
\centering
\caption{\textbf{Model-level WRBench diagnostic profile by viewpoint condition type.} For 23 models, CamPrec is strict local requested-control precision, CamAlign common-yaw intent, Reobs. support the fraction of outputs with judgeable hidden-and-returned evidence, and Reobs. spatial/state conditional means over that subset. The rightmost column \emph{Avg.}\ is the mean of the six metric scores (CamAlign, Integ., the two visible and two re-observed scores), excluding re-observation support. \WRDagger{Daggers} mark sparse support; \WRBest{best}/\WRSecond{second-best} marks are computed within each viewpoint condition type (per column), not across the whole table.}
\label{tab:main-results}

\WRTableTightSetup{2pt}

\begin{tabularx}{\linewidth}{@{}L{0.23\linewidth}YYYYYYYY!{\vrule}Y@{}}
\toprule
\WRTableHead
\WRHeadOne{Model} &
\shortstack{Cam\\Prec.} &
\shortstack{Cam\\Align.} &
\WRHeadOne{Integ.} &
\shortstack{Reobs.\\support} &
\shortstack{Vis.\\spatial} &
\shortstack{Vis.\\state} &
\shortstack{Reobs.\\spatial} &
\shortstack{Reobs.\\state} &
\WRHeadOne{Avg.} \\
\midrule

\GroupHeader{Source-video condition} \\

ReCamMaster & \snd{0.717} & \snd{0.729} & \snd{0.740} & \snd{58.5\%} & \snd{0.715} & \snd{0.535} & \snd{0.665} & \snd{0.616} & \snd{0.667} \\
HyDRA & \fst{0.822} & \fst{0.855} & 0.691 & 33.2\% & 0.648 & 0.500 & 0.509 & 0.445 & 0.608 \\
InSpatio World 14B & 0.693 & 0.661 & \fst{0.824} & \fst{62.3\%} & \fst{0.821} & \fst{0.668} & \fst{0.734} & \fst{0.664} & \fst{0.729} \\

\midrule

\GroupHeader{Geometry-cache condition} \\

Gen3C & 0.699 & \fst{0.764} & 0.749 & \fst{73.0\%} & \snd{0.723} & \fst{0.558} & \fst{0.681} & \fst{0.640} & \fst{0.686} \\
Spatia & \snd{0.704} & 0.482 & \snd{0.763} & 25.8\% & \fst{0.731} & \snd{0.541} & 0.600 & \snd{0.586} & 0.617 \\
VerseCrafter & \fst{0.781} & \snd{0.667} & \fst{0.846} & \snd{28.0\%} & 0.707 & 0.508 & \snd{0.607} & 0.584 & \snd{0.653} \\

\midrule

\GroupHeader{Model-inferred condition} \\

Wan-Fun 2.1-14B & 0.757 & 0.526 & 0.846 & 18.2\% & 0.733 & 0.530 & 0.659 & 0.621 & 0.652 \\
Wan-Fun 2.1-1.3B & \snd{0.771} & \snd{0.729} & 0.842 & 13.8\% & 0.725 & 0.513 & 0.709 & 0.657 & \snd{0.696} \\
Wan-Fun 2.2-5B & 0.724 & 0.335 & 0.812 & 12.0\% & 0.805 & 0.607 & 0.709 & \snd{0.664} & 0.655 \\
Wan-Fun 2.2-A14B & 0.758 & 0.553 & 0.848 & 17.6\% & 0.810 & 0.625 & 0.698 & 0.649 & \fst{0.697} \\
Lingbot World Cam & 0.513 & 0.175 & \fst{0.870} & 6.0\% & \fst{0.876} & \fst{0.735} & \WRSparseSecond{0.717} & \WRSparse{0.663} & 0.673 \\
Lingbot World Act & 0.468 & 0.168 & \snd{0.856} & 6.4\% & \snd{0.874} & \snd{0.719} & \WRSparseBest{0.771} & \WRSparseBest{0.725} & 0.685 \\
LiveWorld & \fst{0.812} & \fst{0.856} & 0.775 & \fst{39.6\%} & 0.703 & 0.541 & 0.661 & 0.600 & 0.689 \\
Hunyuan GameCraft & 0.534 & 0.361 & 0.705 & 6.0\% & 0.672 & 0.440 & \WRSparse{0.554} & \WRSparse{0.490} & 0.537 \\
Hunyuan WorldPlay & 0.708 & 0.261 & \fst{0.870} & \snd{20.8\%} & 0.737 & 0.523 & 0.640 & 0.603 & 0.606 \\
MagicWorld & 0.764 & 0.720 & 0.543 & 15.0\% & 0.623 & 0.458 & 0.584 & 0.574 & 0.584 \\

\midrule

\GroupHeader{Prompt-only condition} \\

Hailuo 2.3 & -- & \snd{0.075} & 0.829 & 6.3\% & \fst{0.891} & \fst{0.759} & \WRSparse{0.719} & \WRSparse{0.642} & \snd{0.652} \\
HappyHorse 1.0 I2V & -- & 0.025 & \snd{0.860} & \snd{10.3\%} & \snd{0.875} & \snd{0.715} & \snd{0.779} & \snd{0.695} & \fst{0.658} \\
Kling v2.6 & -- & \fst{0.094} & \fst{0.864} & 3.3\% & 0.854 & 0.674 & \WRSparse{0.711} & \WRSparse{0.617} & 0.636 \\
Wan2.2 I2V Plus & -- & 0.013 & 0.800 & 2.0\% & 0.829 & 0.644 & \WRSparse{0.714} & \WRSparse{0.610} & 0.602 \\
Wan2.6 I2V & -- & 0.016 & 0.856 & \fst{12.7\%} & 0.855 & 0.682 & 0.659 & 0.556 & 0.604 \\
Wan2.7 I2V & -- & 0.020 & 0.750 & 6.0\% & 0.848 & 0.676 & \WRSparse{0.715} & \WRSparse{0.638} & 0.608 \\
WanX2.1 I2V Turbo & -- & 0.030 & 0.713 & 1.0\% & 0.839 & 0.651 & \WRSparseBest{0.855} & \WRSparseBest{0.777} & 0.644 \\

\bottomrule
\end{tabularx}
\end{table}

\subsection{Diagnostic Analysis: From Common Failures to Model Design Implications}
\label{sec:analysis}

\textbf{Overview.} A persistent world model must bind three things into one shared state: how the camera moves, how the target is displaced, and how an event changes the world. WRBench's most informative failures are \emph{coupling} failures, because a model can render readable frames, plausible camera motion, and a coherent visible action, yet still fail to bring the hidden target back into a judgeable view, or bring it back with the wrong spatial relation or state. We read every model along three axes of increasing depth. (i)~Across \emph{evaluation dimensions} in Section~\ref{sec:metric_dataset_common}, we isolate the failure shared by all 23 models, which is that visible plausibility, re-observation access, and re-observed-state correctness do not move together. (ii)~Across \emph{world-model paradigms} in Section~\ref{sec:model_subtype}, the viewpoint interface reorders which models can even pose the return test, but not whether they pass it. (iii)~Along \emph{model series} in Section~\ref{sec:series_diagnostics}, we trace which increments of scale, architecture, and training actually repair that shared failure. The first two axes establish a model-independent problem, while the third asks what fixes it.

\subsubsection{Common Failures across Evaluation Dimensions}
\label{sec:metric_dataset_common}

\finding{World consistency must be measured on return, not read off frame quality}{Visible-fidelity and re-observation-access scores form their own correlated blocks but predict re-observed-state correctness only weakly, so it is a distinct capability that benchmarks and models must target directly rather than infer from image quality or camera precision.}

\begin{figure}[t]
\centering
\begin{minipage}[t]{0.45\linewidth}
\centering
\includegraphics[width=\linewidth]{\detokenize{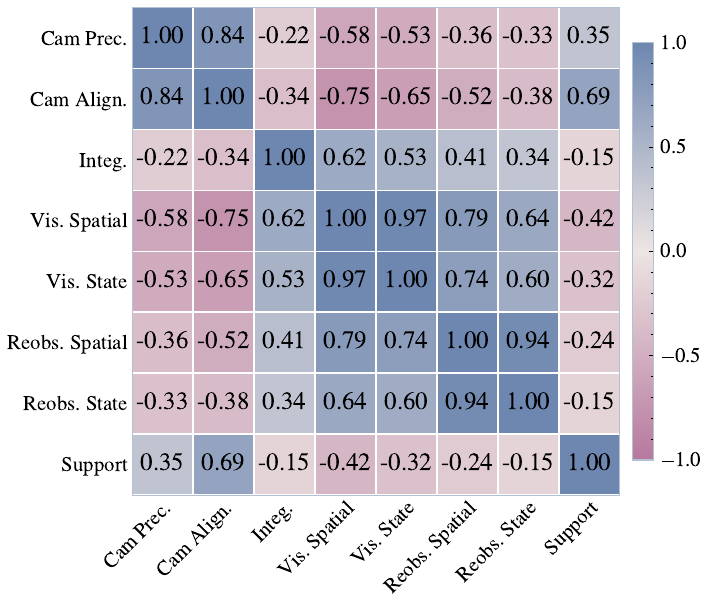}}
\caption{\textbf{Metric correlation structure.} Model-level Pearson correlations among WRBench diagnostic metrics over the 23 evaluated rows. Visible and re-observed consistency form local blocks, while support remains separate from visual integrity.}
\label{fig:analysis-corr-heatmap}
\end{minipage}\hfill
\begin{minipage}[t]{0.53\linewidth}
\centering
\includegraphics[width=0.88\linewidth]{\detokenize{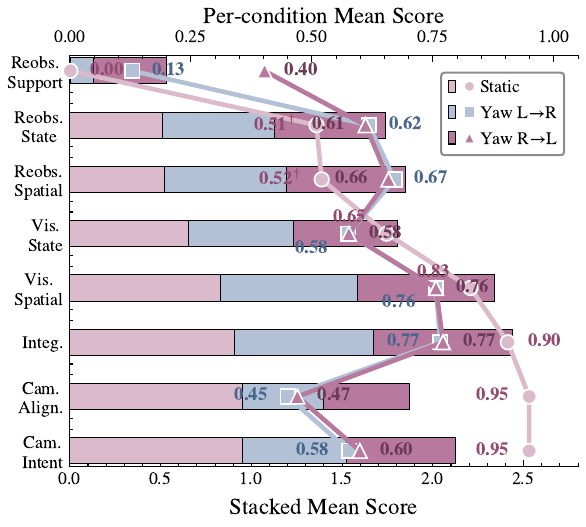}}
\caption{\textbf{Camera motion governs access, not correctness.} Grouped bars compare static, left-to-right yaw, and right-to-left yaw conditions. Camera motion mainly changes re-observation support, while re-observed consistency moves much less.}
\label{fig:analysis-camera-bars}
\end{minipage}
\end{figure}

Visual integrity and camera access predict whether the return question is \emph{asked}, not whether it is answered correctly. A coherent world model would make the diagnostics move as one body, with better rendering buying more returns and a returned object carrying the right position and state. Instead, Figure~\ref{fig:analysis-corr-heatmap} shows three loosely-linked groups. Visible spatial and state lock together at $r{=}0.97$, and the re-observed pair forms its own block at $r{=}0.94$, yet the two blocks track each other only moderately, since visible$\to$re-observed reaches just $r{=}0.60$--$0.79$. Re-observation support, which records whether the object ever leaves and returns in a checkable form, sits apart from all of them and even inverts, falling to $r{=}-0.42$ with visible spatial and to $-0.15$ with both visual integrity and re-observed state.

Each capability a model adds removes one excuse and exposes the next, so the universal failure lives at a single link, the re-observed state, which cleaner images and more access \emph{reach} rather than remove. The cleanest-image systems test the least. The prompt-only APIs and LingBot post the top visible scores, with Hailuo at $0.891/0.759$ and LingBot at $0.876/0.735$, yet they keep the object on screen, so it returns in only $3$--$6\%$ of clips and the hidden-state question is rarely posed. Source-video and geometry-cache interfaces remove that barrier: Gen3C returns the object in $73\%$ of clips, while InSpatio and ReCamMaster do so in $58$--$62\%$. This only surfaces the next failure, because once the object is back and checkable, its re-observed state still scores only $\approx0.62$--$0.66$. Figure~\ref{fig:analysis-camera-bars} isolates the mechanism. Turning the camera swings re-observation support by roughly two orders of magnitude, from near zero under a static camera to ${\sim}40\%$ under yaw, while the re-observed scores barely move; across the two well-supported yaw directions, support triples from $13\%{\to}40\%$ yet re-observed state shifts under $0.01$. Camera motion therefore decides whether the test can be run, not its result. Appendix~\ref{app:forensic-decouple} shows one clip where all three stage-local successes hold yet the returned world state is wrong. The decoupling is sharpest at the extremes, where the model with the best camera execution returns the worst world state while the one with the highest access degrades the scene as the camera moves, as Appendix~\ref{app:forensic-highscore} details.

\finding{World models struggle with object state changes, while spatial changes help them leverage pretrained motion priors to return the object more reliably}{Object \emph{transformation}, not trajectory, is the open problem. Factoring every Natural-25 event into a spatial change where the object is relocated and an in-place state change where it transforms where it sits, e.g.\ folding or tipping, a move hands the model a new coordinate to track and the static scene to anchor against, so the object returns roughly right, whereas an in-place change offers no such anchor and leaves the altered object to drift and smear on return.}

\begin{figure}[t]
\centering
\includegraphics[width=\linewidth]{\detokenize{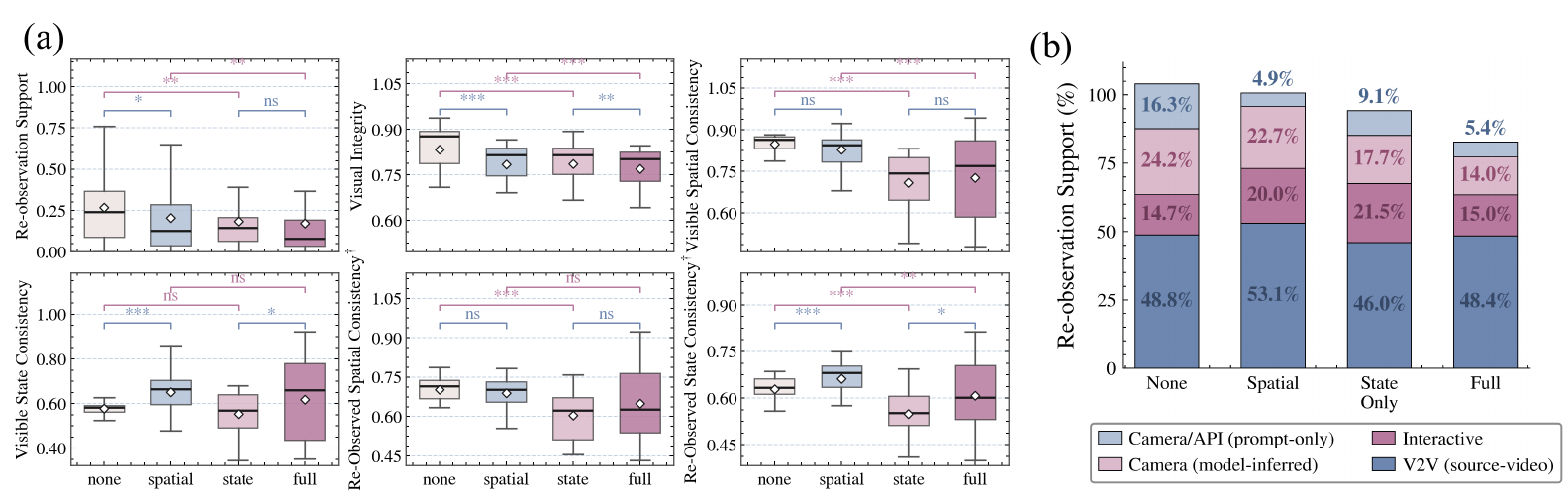}}
\caption{\textbf{Event-factor pressure splits by factor.} Boxplots show per-model metric distributions under the Natural-25 $2\times2$ event design; bars summarize event-factor pressure on out-of-view behavior. Contrasts separate relocation from in-place state change; stars denote paired Wilcoxon tests (* $p<0.05$, ** $p<0.01$, *** $p<0.001$; n.s. not significant).}
\label{fig:analysis-event-factor}
\end{figure}

Moving an object and changing it in place pull in opposite directions. Having localized the failure to the re-observed link, we ask which events break it. Figure~\ref{fig:analysis-event-factor} factors every Natural-25 event into two independent axes, whether the object moves and whether its state changes, toggling each while holding the other fixed. Adding a move barely touches the visible position score, at $+0.008$ (n.s.), while lifting the later scores: visible state rises $+0.070$ and re-observed state $+0.038$ at paired Wilcoxon $p<0.01$. Adding an in-place change does the reverse, and the tell is \emph{where} it lands. It depresses the visible \emph{position} score the most, at $-0.114$ ($p<0.001$), while leaving the visible state score essentially flat at $-0.031$ (n.s.), and it drags down both re-observed scores, re-observed spatial $-0.075$ ($p<0.01$) and re-observed state $-0.068$ ($p<0.001$). The position hit is the mechanism: a relocation gives the model a new coordinate to track and the static scene to anchor against, whereas an in-place change offers no new location, so the altered object drifts and smears where it sits. The damage stays local, since overall visual integrity falls only $-0.029$, so the frame holds while the changed object degrades.

Two checks rule out the easy reading. The effect is not just rarer returns. At matched return rates, in right-to-left yaw where moved and in-place events return at $40.5\%$ vs.\ $40.4\%$, in-place events still score lowest on visible position and state, at $0.661/0.502$ vs.\ $0.819/0.629$. Nor do the axes interact: the per-model interaction term on re-observed state is indistinguishable from zero, with mean $-0.009$, Wilcoxon $p=0.45$, $n=13$, so a move adds its benefit on top of an in-place change rather than rescuing it. The single hardest case for today's video world models is thus a sharp one, preserving an object's changed state across a viewpoint change when the change moved it nowhere. Appendix~\ref{app:forensic-inplace} traces it frame by frame, showing that in a fixed scene the changed object returns in the wrong state, is silently erased, or is lost during occlusion, while the room behind it stays sharp.

These two failures, a re-observed-state link that neither quality nor access ever closes and an in-place hard case that no event design escapes, are common to all 23 models and independent of architecture or interface. The next two subsections ask what \emph{does} move them, namely the world-model paradigm through which a model is told where to look in Section~\ref{sec:model_subtype}, and the scale, architecture, and training increments within a model series in Section~\ref{sec:series_diagnostics}.

\subsubsection{World-Model Paradigms by Viewpoint Condition Type}
\label{sec:model_subtype}
\label{sec:equiv}

\finding{World-model controllability has mastered static re-rendering, not dynamic evolution}{Access rises with how much already-seen footage a paradigm carries, but every paradigm only replays the recorded scene, so the next controllability target is off-screen \emph{dynamics}, not just viewpoint.}

We read the four viewpoint condition types as \emph{controllability} paradigms rather than individual designs, where each paradigm fixes what a model can control about an already-seen scene: a prompt-only condition carries a sentence, a model-inferred condition carries only internal camera, action, or state controls, a source-video condition carries a reference stream of appearance and dynamics, and a geometry-cache condition carries a point-cloud, 3D, or 4D record of the scene. The right panel of Figure~\ref{fig:wrbench} previews how these paradigms compare, plotting the best value each condition type reaches on every dimension. The access and visible-quality frontiers separate sharply by paradigm, while the re-observed-consistency frontier rises far less and, where it does look high, rests on sparse re-observation. A richer paradigm therefore mostly buys access and readable frames rather than a correct return. The paradigms differ in one thing, how much external record of the already-seen scene they externalize, and the two questions from Section~\ref{sec:metric_dataset_common}, whether the hidden object returns and whether it returns in the right state, separate cleanly along that single axis.

Table~\ref{tab:gen3c-baseline} ranks representative paradigms by how often they re-expose the hidden target, and access falls monotonically down the ladder as the external record thins. Prompt-only conditions sit at the floor, with Hailuo at $6.3\%$ and Kling at $3.3\%$: given only a sentence, the camera moves conservatively and keeps the target framed, so a hidden-then-returned test is rarely even created. A model-inferred condition raises access, since Wan-Fun A14B reaches $17.6\%$ while synthesizing the new view internally, though at lower visible consistency than the prompt-only systems, $0.718$ versus $0.825$. Source-video and geometry-cache conditions top the ladder, with Gen3C at $73\%$, InSpatio at $62\%$, ReCamMaster at $58\%$, and VerseCrafter at $28\%$, because an external record can be \emph{replayed} or \emph{re-projected} to put the target back where it was last seen. The amount of already-seen footage carried, rather than the cache label, sets access. Gen3C derives its 3D cache from a source video and so behaves as V2V and tops the ladder, whereas the text- and image-conditioned (TI2V) geometry caches such as VerseCrafter and Spatia carry less and sit well below the source-video rows. Access is therefore a property of the paradigm rather than of any one design, since handing the model a record of the scene, in whatever form, is what lets the question be posed at all. Appendix~\ref{app:forensic-access} shows the two ends of this spectrum frame by frame, and Appendix~\ref{app:condition-detail-notes} details what each paradigm externalizes.

\begin{table}[t]
\centering
\begin{minipage}[b]{0.496\linewidth}
\centering
\caption{\textbf{Condition-interface access profile.} Representative interfaces ranked by Reobs.\ support; Subtype follows the viewpoint-condition taxonomy; Cam Align.\ is common-yaw alignment and Vis.\ avg.\ mean visible consistency.}
\label{tab:gen3c-baseline}
\WRTableTightSetup{0.55pt}\footnotesize\renewcommand{\arraystretch}{1.10}
\begin{tabularx}{\linewidth}{@{}M{0.28\linewidth}C{0.295\linewidth}YYY@{}}
\toprule
\WRTableHead
\WRHeadOne{Model} & \WRHeadOne{Subtype} & \shortstack{Reobs.\\support} & \shortstack{Cam\\Align.} & \shortstack{Vis.\\avg.} \\
\midrule
Gen3C        & \WRCondGeom   & 73.0\% & 0.764 & 0.641 \\
InSpatio~14B & \WRCondSource & 62.3\% & 0.661 & 0.745 \\
ReCamMaster  & \WRCondSource & 58.5\% & 0.729 & 0.625 \\
VerseCrafter & \WRCondGeom   & 28.0\% & 0.667 & 0.608 \\
Wan-Fun~A14B & \WRCondModel  & 17.6\% & 0.553 & 0.718 \\
Hailuo~2.3   & \WRCondPrompt &  6.3\% & 0.075 & 0.825 \\
Kling~v2.6   & \WRCondPrompt &  3.3\% & 0.094 & 0.764 \\
\bottomrule
\end{tabularx}
\end{minipage}\hfill
\begin{minipage}[b]{0.496\linewidth}
\centering
\caption{\textbf{Re-observed-state event gap.} Representative interfaces with re-observed state on relocation vs.\ in-place events; Subtype follows the viewpoint-condition taxonomy, and Gap is relocation minus in-place, the in-place penalty.}
\label{tab:analysis-condition-event}
\WRTableTightSetup{0.55pt}\footnotesize\renewcommand{\arraystretch}{1.10}
\begin{tabularx}{\linewidth}{@{}M{0.27\linewidth}C{0.30\linewidth}YYC{0.15\linewidth}@{}}
\toprule
\WRTableHead
\WRHeadOne{Model} & \WRHeadOne{Subtype} & \shortstack{Spatial\\event} & \shortstack{State\\event} & \WRHeadOne{Gap} \\
\midrule
Gen3C        & \WRCondGeom   & 0.711 & 0.559 & \WRGapSig{0.152} \\
InSpatio~14B & \WRCondSource & 0.720 & 0.591 & \WRGapSig{0.129} \\
Spatia       & \WRCondGeom   & 0.633 & 0.512 & \WRGapSig{0.121} \\
VerseCrafter & \WRCondGeom   & 0.638 & 0.527 & \WRGapSig{0.111} \\
ReCamMaster  & \WRCondSource & 0.686 & 0.589 & \WRGapSig{0.097} \\
HyDRA        & \WRCondSource & 0.475 & 0.414 & \WRGapSig{0.061} \\
Wan-Fun~A14B & \WRCondModel  & 0.682 & 0.611 & \WRGapSig{0.071} \\
\bottomrule
\end{tabularx}
\end{minipage}
\end{table}

Re-observed-state correctness, however, does not track that ladder. Table~\ref{tab:analysis-condition-event} reports re-observed state on the two event families from Section~\ref{sec:metric_dataset_common}, where in-place change trails relocation on every paradigm. On the geometry-cache and source-video rows that lead on access, the shortfall is a consistent $\approx0.10$--$0.15$, with Gen3C at $0.711$ versus $0.559$, InSpatio at $0.720$ versus $0.591$, and Spatia at $0.633$ versus $0.512$, exactly the universal hard case the prior section isolated. The ordering \emph{inside} that gap is the tell: the paradigm best at re-projecting a static scene, the geometry cache in Gen3C, opens the \emph{largest} gap of all at $0.152$, since a stored record of where surfaces were cannot reconstruct a change it never observed. The decoupling is sharpest read across paradigms. Moving from the model-inferred to the geometry-cache paradigm more than quadruples access, from Wan-Fun A14B at $17.6\%$ to Gen3C at $73\%$, yet leaves relocation re-observed state essentially flat, from $0.682$ to $0.711$, and pushes in-place re-observed state \emph{down}, from $0.611$ to $0.559$. A richer paradigm therefore only moves the bottleneck from \emph{whether the object returns} to \emph{whether the return reflects an unobserved change}. This second wall is identical across all four paradigms, because it is the in-place collapse of Section~\ref{sec:metric_dataset_common} re-expressed along the model axis: the paradigm reorders which models can pose the test but not the ceiling on answering it. Appendix~\ref{app:forensic-crossinterface} runs one in-place fold through all four paradigms and finds four distinct failure signatures, none recovering the unobserved change.

\subsubsection{Series Increments: Scaling, Architecture, and Training}
\label{sec:series_diagnostics}

\begin{figure}[t]
\centering
\begin{minipage}[t]{0.315\linewidth}
\centering
\includegraphics[width=\linewidth]{\detokenize{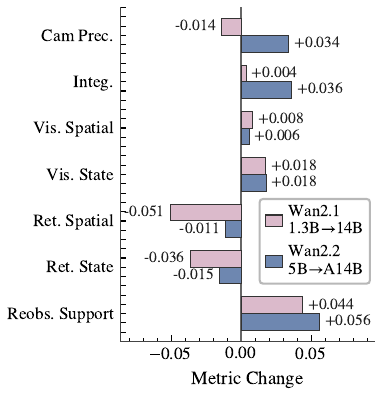}}
\end{minipage}\hfill
\begin{minipage}[t]{0.378\linewidth}
\centering
\includegraphics[width=\linewidth]{\detokenize{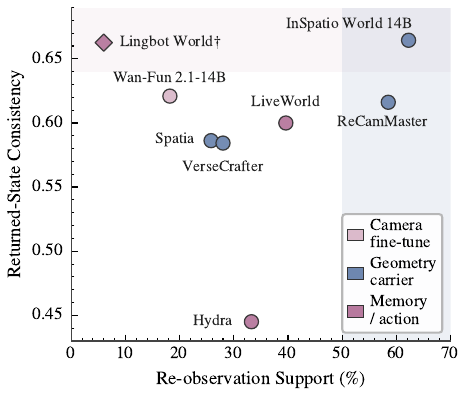}}
\end{minipage}\hfill
\begin{minipage}[t]{0.285\linewidth}
\centering
\includegraphics[width=\linewidth]{\detokenize{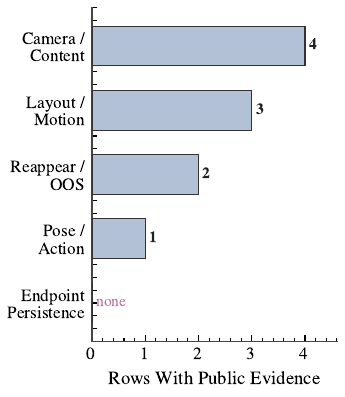}}
\end{minipage}
\caption{\textbf{Series diagnostics.} Panels summarize scale/version, architecture, and training-signal diagnostics for Wan-derived rows. The common pattern is that access improves more readily than endpoint persistence.}
\label{fig:series-diagnostics}
\end{figure}

\noindent\textbf{Series diagnostics as a world-model stress map.}
Across all twelve Wan-derived rows and all three axes, the same pattern holds: access improves more readily than endpoint persistence, so the in-place change that prior sections isolated as the universal hard case stays the unbound endpoint that no increment yet reaches. We treat the twelve rows as one controlled stress map over a shared video prior rather than a capacity leaderboard, because each row shares a documented Wan base but varies one increment while holding the family roughly fixed. Figure~\ref{fig:series-diagnostics} reads that map along three axes, scale and version in panel~a, architecture design in panel~b, and training signal in panel~c, with per-row mechanism coding in Table~\ref{tab:appendix-wan-increment-taxonomy} and Appendix~\ref{app:series-comparison-notes}. Each axis asks the endpoint question from Sections~\ref{sec:metric_dataset_common}--\ref{sec:model_subtype}: when an event resolves out of view and the camera later returns, does the increment only make that moment easier to see and score, which is re-observation support, or does it also preserve how the event resolved, which is re-observed spatial and re-observed state consistency, conditional over the judgeable re-observation subset?

\finding{Regular video-supervised scaling will not deliver world-state evolution, so it has to be designed in}{
Larger Wan backbones add re-observation access while the 2.1$\to$2.2 upgrade adds visible fidelity, which are the observable axes such training already optimizes, yet conditional re-observed state, scored only on the clips where the object is re-observed, stays in a fixed band, so endpoint persistence is a target for architecture and training rather than for more parameters or standard video data.
}

Scaling and version upgrades buy observable quality but not endpoint persistence, because modern video scaling is tuned to \emph{observable} axes. Wan's own reports frame scaling as improving motion quality, visual fidelity, camera control, and instruction adherence: $1.3$B${\to}14$B lifts the Wan-Bench weighted score $0.689{\to}0.724$ and the VBench total $83.96\%{\to}86.22\%$, while Wan~2.2 explicitly targets cinematic aesthetics and complex motion~\citep{wan2025wan}. All of these are scored from generated pixels and text alignment, never from hidden-state persistence after occlusion~\citep{huang2024vbench}. WRBench measures the orthogonal axis. The only clean within-family scale pairs are the four Wan-Fun rows in Figure~\ref{fig:series-diagnostics}a. Wan~2.1 $1.3$B${\to}14$B raises re-observation support $13.8\%{\to}18.2\%$ but lowers re-observed state consistency $0.657{\to}0.621$, and Wan~2.2 $5$B${\to}$A14B raises support $12.0\%{\to}17.6\%$ while re-observed state slips $0.664{\to}0.649$, even though A14B does improve requested-camera precision, visual integrity, and visible consistency over 5B. A version axis sits inside the same rows: the Wan~2.2 variants reach markedly higher visible spatial and state consistency, at $0.805/0.607$ and $0.810/0.625$, than either Wan~2.1 row at $0.725/0.513$ and $0.733/0.530$, yet re-observed state never leaves the $0.621$--$0.664$ band. Read against the other eight rows, where source-video and geometry carriers span $25.8$--$62.3\%$ support while LingBot enters the judgeable subset only $6.0$--$6.4\%$ of the time, the shortfall is compositional rather than a scaling law: a stronger prior surfaces more returns and cleaner frames without preserving what happened out of view, which is exactly the in-place hard case of Sections~\ref{sec:metric_dataset_common}--\ref{sec:model_subtype}. Scaling and version claims should therefore report re-observation support and re-observed state consistency separately, and A14B denotes active-14B MoE context rather than dense parameter scaling, with the full per-row map in Appendix~\ref{app:series-comparison-notes}.

\finding{World models need a \emph{what}-memory, not just a \emph{where}-memory}{
Every architecture caches geometry, appearance, or motion to re-render \emph{where} the scene was, so how the camera is \emph{encoded} is second-order, while the missing, highest-leverage component is a state writer that records \emph{what} changed while hidden.
}

The state carrier, not the camera-conditioning format, is the architectural bottleneck: reading the full roster as a ladder shows no carrier closes the endpoint (Figure~\ref{fig:series-diagnostics}b). Two architectural decisions sit behind each camera-controllable row. The first is the camera-conditioning representation, whether Pl\"ucker-coordinate ray embeddings as in CameraCtrl~\citep{he2024cameractrl}, discretized or one-hot pose tokens, or direct injection of the camera-to-world (C2W) extrinsic matrix as in MotionCtrl~\citep{wang2024motionctrl}. This choice is second-order for WRBench. Camera-control rows cluster on access and requested-camera precision regardless of encoding, and within-method ablations report only a small representation effect next to the access-and-endpoint gap measured here, so how the camera is \emph{encoded} is not the bottleneck. The decision that moves WRBench is the \emph{state carrier}, which never closes as the roster climbs. Camera-lens control alone, the four Wan-Fun rows at $12$--$18\%$ support, stores nothing out of view; external geometry adapters such as VerseCrafter's GeoAdapter and Spatia's point-cloud memory, at $26$--$28\%$, add a static layout record yet sit near the series floor on re-observed state at $D_6\approx0.58$; source-video carriers such as ReCamMaster and InSpatio, at $58$--$62\%$, stream appearance and dynamics to the highest access but borrow them from the condition at $D_6\approx0.62$--$0.66$; explicit reappearance memory, HyDRA's Memory Tokenizer and Dynamic Retrieval Attention, wins the highest requested-camera precision at $0.822$ yet the weakest re-observed state of any row at $0.445$; a dedicated state adapter, LiveWorld at $39.6\%$, lifts access without closing the endpoint at $0.600$; and pose/action conditioning in LingBot preserves the strongest visible state of any Wan-derived row but lets support collapse to $6\%$. Each rung repairs the previous weak link, yet every carrier stores \emph{where} to look back rather than \emph{what} changed while hidden. LiveWorld's own \emph{w/o Event Evo} ablation isolates the gap: removing its generative evolution engine collapses foreground fidelity while background geometry stays intact, which confirms that a spatial memory alone never writes hidden dynamics. Read by how each carrier fails when forced to render the unobserved, the modes are distinct but equivalent in outcome. A geometry or source cache replays where the scene was, a generative-forward simulator such as LiveWorld hallucinates the subject forward and returns it duplicated, and a memory trained on synthetic exit and entry data such as HyDRA reverts to its training domain, as Appendix~\ref{app:forensic-highscore} shows. Architecture progress should therefore target writing the event endpoint into state rather than the camera-encoding format, with the full per-row reading in Appendix~\ref{app:series-comparison-notes}.

\finding{Endpoint persistence is an unwritten training objective, so write it with a long-to-short recipe}{
No \emph{public} loss or documented post-training stage supervises the unobserved outcome, so we propose a \emph{long-to-short} recipe: learn persistence on long horizons, then add explicit camera-execution supervision.
}

No public training signal writes the unobserved endpoint into state, so reading the public training ladder shows exactly where the gradient stops, as Figure~\ref{fig:series-diagnostics}c records, treating unreported recipes as evidence gaps rather than negative evidence. Camera-control fine-tuning on the Wan-Fun rows, whose public scope and control data are unreported, supplies lens conditioning only, so support stays low at $12.0$--$18.2\%$ and scaling expands judgeable returns without improving re-observed state. ReCamMaster~\citep{bai2025recammaster} adds the first explicit disentanglement signal, where MultiCamVideo pairs one event across ten synchronized cameras to separate viewpoint motion from scene content, so support jumps to $58.5\%$, yet re-observed state stays $0.616$ because no label writes the hidden endpoint. Geometry and layout training, including VerseCrafter's frozen-backbone GeoAdapter on rendered 4D maps, Spatia's point-cloud memory, and InSpatio's depth-guided source path, raises access to $25.8$--$62.3\%$ while re-observed state holds at $0.58$--$0.66$, because external structure carries \emph{where} to look rather than \emph{what} changed. Reappearance and state-channel training, such as HyDRA's HM-World exit and entry events and LiveWorld's out-of-sight adapter, moves retrieval and latent dynamics without binding the endpoint, since HyDRA reaches the top requested-camera precision but the weakest re-observed state at $0.445$. LingBot~\citep{team2026advancing} inverts the ladder. Its long-trajectory world-model training holds the visible world state best of any Wan-derived row at $0.876/0.735$, yet the same objective binds the camera to object dynamics, so requested displacement barely executes and support collapses to $6.0$--$6.4\%$ with returned rows $n{=}30/32$. This leaves the persisted state unjudgeable exactly when WRBench would check it. We read this as a reasoned proposal rather than a demonstrated result: a \emph{long-to-short} strategy that learns persistence from long horizons, then adds a shorter, controllable camera-execution objective with explicit requested-displacement supervision, so that the preserved state becomes re-observable on return. Endpoint-directed reward or policy training remains only proposed in Appendix~\ref{app:preference-pair-export}, and the target is the in-place endpoint of Sections~\ref{sec:metric_dataset_common}--\ref{sec:model_subtype}, an outcome that must survive non-observation and that no current public signal writes back into state. Crucially, no documented stage is reinforcement learning on the endpoint. Every post-training recipe here is distribution-matching distillation or reward-weighted DMD, and the synthetic paired data that buys the strongest access can revert to its training domain when asked for the unobserved, as Appendices~\ref{app:series-comparison-notes} and~\ref{app:forensic-highscore} show.

\subsection{Validating Human Alignment of WRBench}
\label{sec:human_alignment_analysis}

\noindent\textbf{Human calibration grounds the diagnostic profile.}
The human study validates WRBench dimensions as dimension-specific comparisons rather than collapsing them into a single preference label. Agreement is strong enough to support the main claims about requested-view execution, visual integrity, re-observation support, and conditional re-observed consistency.

\begin{wraptable}{r}{0.50\linewidth}
\vspace{-0.35cm}
\centering
\WRWrapTableSetup{1.8pt}
\caption{\textbf{Human calibration snapshot.} Agree. is AC1 on the main reliability set, Bridge $N$ counts pairs where an evaluator--human bridge exists, and Rev. counts opposite-direction threshold decisions; visual integrity uses a separate 190-pair holdout.}
\begin{tabularx}{\linewidth}{@{}Xcccc@{}}
\toprule
\WRTableHead
\WRHeadOne{Metric} & \shortstack{Agree.\\(AC1)} & \shortstack{Rank\\$\rho$} & \shortstack{Bridge\\$N$} & \shortstack{Rev.\\$n/N$} \\
\midrule
View exec. & 0.898 & -- & -- & -- \\
Visual integrity & 0.877 & 0.709 & 190 & -- \\
\focusrow Vis. spatial & 0.788 & 0.386 & 228 & 0/228 \\
Vis. event-state & 0.790 & 0.484 & 229 & 0/229 \\
Reobs. spatial & 0.875 & 0.667 & 136 & 1/136 \\
Reobs. event-state & 0.937 & 0.660 & 136 & 8/136 \\
\bottomrule
\end{tabularx}
\label{tab:human-calibration-snapshot}
\vspace{-0.30cm}
\end{wraptable}

Table~\ref{tab:human-calibration-snapshot} shows the calibration pattern. AC1 is high for requested-view execution, visual integrity, re-observed spatial consistency, and re-observed event-state consistency. Opposite-direction threshold decisions are rare: none for visible spatial and visible event-state, one of 136 for re-observed spatial consistency, and eight of 136 for re-observed event-state consistency. Human disagreement mainly appears as ties around small differences rather than as systematic reversal, which is exactly the pattern expected from a benchmark that separates nearby diagnostic states.

This calibration is especially important for the access--re-observed-consistency separation. A video that never hides the target, or never returns identifiable evidence, cannot support re-observed-state evaluation. Once a clip does provide judgeable hidden-and-returned evidence, re-observed spatial and re-observed state consistency evaluate the endpoint itself. Human-aligned judgeability therefore makes the denominator meaningful and keeps access distinct from correctness.

The same design supports model comparison and future reward use. Sparse re-observed rows are flagged because they describe a small judgeable subset, while re-observed metrics are interpreted only on clips where the hidden-and-returned evidence exists. Preference or reward mining can use this structure directly: reward visible plausibility, camera access, and re-observed endpoint consistency as separate targets, and avoid treating a clip that skips the re-observed-state test as evidence of successful evolution consistency.

\section{Conclusion}
\label{sec:conclusion}

We introduced WRBench, a diagnostic benchmark that evaluates whether video generation models maintain evolving world state under camera-induced partial observation. Across 23 models and 9,600 generated outputs, WRBench reveals a preservation--access--re-observed-consistency gap: models can preserve visible evidence, execute plausible camera motion, or expose the target region again, but they do not reliably maintain the event endpoint after re-observation. The shared bottleneck is endpoint binding, because a persistent world model should not merely render a plausible returned view but should write the event-induced object relation, contact endpoint, posture, containment, or collision endpoint back into the returned scene. By separating requested-camera precision, prompt-camera alignment, visual integrity, visible spatial/state consistency, re-observation support, and re-observed spatial/state consistency, WRBench provides a compact diagnostic profile that evaluates future systems as persistent world models rather than view-conditioned renderers that master static re-rendering but not dynamic world evolution.

\newpage
\bibliographystyle{unsrtnat}
\bibliography{iclr2026_conference}

\newpage
\appendix

\phantomsection
\section*{Appendix}
\label{appendix}
\paragraph{Table of contents:}
\begin{itemize}
  \item \S\ref{appendix:use_of_LLM}: Use of Large Language Models (LLMs)
  \item \S\ref{app:repro}: Reproducibility and Metric--Dataset Records
  \item \S\ref{app:limitations_future}: Limitations and Future Work
  \item \S\ref{app:metric-common-problems}: Metric--Dataset Common-Problem Checks
  \item \S\ref{app:visual-integrity-implementation}: Visual-Integrity Implementation
  \item \S\ref{app:forensics}: Frame-Level Failure Forensics
  \item \S\ref{app:model-subtype-series}: Model Subtype and Series Reading Guide
  \item \S\ref{app:human-alignment-notes}: Human Validation Notes
  \item \S\ref{app:preference-pair-export}: Preference-Pair Export and Reward/Policy Outlook
  \item \S\ref{app:dense-control-future}: Dense-Control and Future Benchmark Extensions
\end{itemize}

\section{Use of Large Language Models (LLMs)}
\label{appendix:use_of_LLM}

Large language models were used as editorial and coding assistants for manuscript organization, wording, LaTeX editing, and consistency checks. All scientific claims, metric definitions, numerical results, tables, figures, and citations remain the responsibility of the authors. LLM assistance was not used as a substitute for benchmark generation, measurement records, human annotation, or empirical validation; any LLM-suggested prose was reviewed against the WRBench evidence dimensions and the paper's denominator conventions before inclusion.

\section{Reproducibility and Metric--Dataset Records}
\label{app:repro}
The main tables are generated after fixing the prompt set, model roster, metric definitions, and aggregation scripts.

\begin{itemize}
  \item Benchmark prompts and first frames include initial-state, event, object/action, and camera-trajectory metadata.
  \item The model roster records model identifiers, conditioning interface, generation settings, and reference-video conditioning for V2V methods.
  \item CamPrec, visual integrity, visible-frame metrics, re-observed-state consistency, and re-observation records are exported before aggregation.
  \item Aggregation scripts generate model, family, camera-protocol, reasoning-tier, and object/action slices.
  \item The human-validation protocol records sampling, annotation instructions, agreement computation, and calibration examples for ambiguous re-observation cases.
\end{itemize}

\section{Limitations and Future Work}
\label{app:limitations_future}

WRBench targets viewpoint-conditioned dynamic world-state attribution: whether camera motion, target-relative displacement, spatial relations, and event endpoints stay bound when observability changes. Natural-25 and the evaluation dimensions are designed to make that endpoint-binding problem measurable. The next step is to use these diagnostics as a model-design loop: train systems with explicit state carriers, mine reward or policy signals from axis-level records, and extend the benchmark with finer VLM-labeled masks, boxes, and dense-control settings. Dense controls should enter with leakage-aware input policies, because target-view depth, segmentation, edge maps, or endpoint controls can reveal the post-event state and turn a world-state test into control following.

\section{Metric--Dataset Common-Problem Checks}
\label{app:metric-common-problems}

The main text analyzes WRBench through diagnostic metrics before comparing viewpoint condition types. Table~\ref{tab:metric-common-problems} records the intended common-problem interpretation for each metric. The same generated-video dataset is therefore not collapsed into one score: each denominator answers a different question about where a world-state claim becomes unsupported.

\begin{table}[t]
\centering
\caption{\textbf{Metric common-problem checks.} Rows state each metric's denominator, diagnostic use, and diagnostic reading.}
\label{tab:metric-common-problems}
\WRTableSetup
\begin{tabularx}{\linewidth}{@{}L{0.19\linewidth} L{0.19\linewidth} L{0.27\linewidth} X@{}}
\toprule
\WRTableHead
Metric & Denominator & Diagnoses & Not for \\
\midrule
Strict CamPrec & 7,500 local rows & Requested trajectory execution & Prompt-only API rows \\
Common-yaw CamAlign & 7,800 yaw rows & Shared yaw intent & Full trajectory precision \\
Visual integrity & 9,600 outputs & Readable visual evidence & Event or physics correctness \\
Visible spatial & 9,600 outputs & Observable spatial errors & Hidden-state consistency \\
Visible state & 9,600 outputs & Observable event-state errors & Hidden-state consistency \\
Re-observation support & 9,600 outputs & Judgeable re-observation access & State correctness \\
Re-observed spatial & 2,073 judgeable rows & Re-observed spatial consistency & Full-denominator persistence \\
Re-observed state & 2,073 judgeable rows & Re-observed event-state consistency & Full-denominator persistence \\
\bottomrule
\end{tabularx}
\end{table}

\subsection{Visual-Integrity Implementation}
\label{app:visual-integrity-implementation}

The visual-integrity score uses the fixed DINOv2 proxy described in Section~\ref{sec:evaluation_method_suite}. For each generated video $v$, we sample frames with a time-based sampler at 3 fps, retain the first and last frames, and cap the sequence at 24 frames. Each sampled RGB frame is resized while preserving the full field of view and padded to the DINOv2 patch grid before standard rescaling and ImageNet normalization. We avoid center cropping because WRBench scenes often place prompt-critical subjects, targets, or failure regions away from the image center. With patch size 14, the processed frame yields one CLS token and a rectangular grid of patch tokens; patches that contain only padding are masked out for local matching.

Let $g_t$ denote the $\ell_2$-normalized DINOv2 CLS token for sampled frame $t$, let $p_{t,i}$ denote its normalized patch token at grid index $i$, and let $\Omega_t$ be the set of non-padding patch indices. The global component measures long-range frame-level appearance continuity over the sampled clip:
\begin{equation}
  s_{\mathrm{global}}(v)
  =
  \left[g_1^\top g_T\right]_{0}^{1},
\end{equation}
where $[\cdot]_{0}^{1}$ clips cosine similarity to $[0,1]$. The local component measures whether local semantic structures can still be matched across adjacent sampled frames. For adjacent frames $t$ and $t+1$, we compute patch-token cosine similarities only over valid patches, $m_{ij}^{(t)}=[p_{t,i}^{\top}p_{t+1,j}]_{0}^{1}$ for $i\in\Omega_t$ and $j\in\Omega_{t+1}$, and form a bidirectional best-match set
\begin{equation}
  \mathcal{B}_t
  =
  \left\{\max_{j\in\Omega_{t+1}} m_{ij}^{(t)}\right\}_{i\in\Omega_t}
  \cup
  \left\{\max_{i\in\Omega_t} m_{ij}^{(t)}\right\}_{j\in\Omega_{t+1}} .
\end{equation}
The adjacent-frame local score is the 20th percentile of $\mathcal{B}_t$, and the video-level local score is again the 20th percentile over adjacent-frame local scores:
\begin{equation}
  s_{\mathrm{local}}(v)
  =
  \operatorname{P}_{20}
  \left(
  \left\{
  \operatorname{P}_{20}(\mathcal{B}_t)
  \right\}_{t=1}^{T-1}
  \right).
\end{equation}
This low-tail aggregation is a heuristic that makes the score sensitive to localized collapse, ghosting, disappearance, hard cuts, or identity drift even when the median frame pair remains visually similar. The best-match formulation does not require a patch to match the same image coordinate, so it is more tolerant to object and camera motion than fixed-location patch cosine. It is still not an object detector, object tracker, or prompt-grounded subject mask.

The global term uses the DINOv2 CLS token rather than a valid-patch mean. Padding patches are excluded from local patch matching, but the CLS token can in principle attend to padded regions. We therefore include a valid-patch-mean global alternative as a sensitivity check. On the adjudicated visual-integrity bridge slice, the CLS-based headline candidate has higher exact agreement, tie accuracy, and weighted $\kappa$ than the valid-patch-mean alternative, so the patch-mean global descriptor is reported only as a diagnostic and is not used for the headline visual-integrity score. The score serves as visual-integrity evidence rather than object-level state tracking or event correctness.

\subsection{Additional Common-Problem Tables and Case Montages}
\label{app:metric-common-extra}

To keep \S\ref{sec:metric_dataset_common} focused on the two central claims, this appendix expands the same slices at finer granularity. Table~\ref{tab:analysis-high-support-re-observed-consistency} separates re-observation support from re-observed consistency, Table~\ref{tab:yaw_retention} summarizes directional access, and Tables~\ref{tab:analysis-event-delta} and~\ref{tab:analysis-action-pressure} expand the event/action diagnostics.

\begin{table}[t]
\centering
\caption{\textbf{Re-observed consistency for high-re-observation-support models.} Rows list re-observation support and conditional re-observed consistency.}
\label{tab:analysis-high-support-re-observed-consistency}
\WRTableAnalysisSetup
\begin{tabular}{llrrr}
\toprule
\WRTableHead
Model / group & Condition type & \shortstack{Reobs.\\support} & Reobs. spatial & Reobs. state \\
\midrule
Gen3C              & Geometry-cache    & 73.0\% & 0.681 & 0.640 \\
InSpatio World 14B & Source-video       & 62.3\% & 0.734 & 0.664 \\
ReCamMaster        & Source-video       & 58.5\% & 0.665 & 0.616 \\
LiveWorld          & Model-inferred     & 39.6\% & 0.661 & 0.600 \\
HyDRA              & Source-video       & 33.2\% & 0.509 & 0.445 \\
\bottomrule
\end{tabular}
\end{table}

Family-level slices follow the same reading and are therefore summarized rather than tabled: higher re-observation support changes what can be judged, but it does not make re-observed consistency automatic; state-only pressure appears as an event-design effect rather than a family leaderboard.

\begin{figure}[t]
\centering
\includegraphics[width=\linewidth]{\detokenize{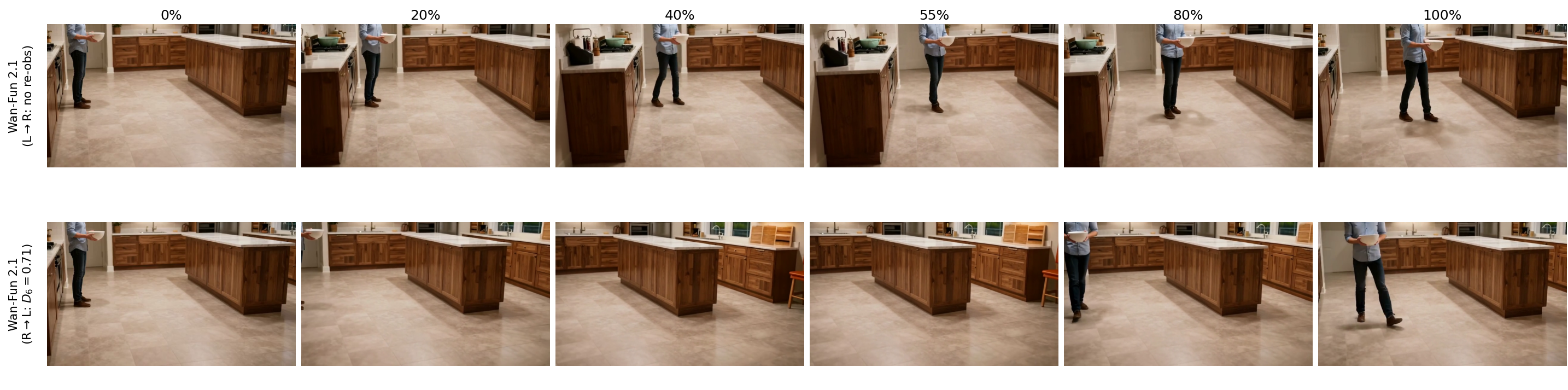}}
\caption{\textbf{Directional re-observation-support contrast (Wan-Fun 2.1).} A kitchen bowl-placement task (spatial event) under two yaw directions. \emph{Top (L$\rightarrow$R):} the subject is dragged with the camera and never exits, so no return is created and re-observed consistency is undefined. \emph{Bottom (R$\rightarrow$L):} the subject leaves and re-enters, creating judgeable re-observation ($D_6=0.71$). Visible spatial/state scores stay similar, so only the camera channel decides whether returned evidence exists.}
\label{fig:analysis-direction-support-case}
\end{figure}

\begin{table}[t]
\centering
\caption{\textbf{Event-change pressure on diagnostic metrics.} Read as a $2\times2$ design: \texttt{none} toggles neither factor, \texttt{spatial} only the spatial-displacement factor, \texttt{state-only} only the state-change factor, and \texttt{full} both.}
\label{tab:analysis-event-delta}
\WRTableAnalysisSetup
\begin{tabular}{lrrrrrr}
\toprule
\WRTableHead
Event change & \shortstack{Reobs.\\support} & Integ. & Vis. spatial & Vis. state & Reobs. spatial & Reobs. state \\
\midrule
none       & 24.2\% & 0.826 & 0.820 & 0.580 & 0.678 & 0.618 \\
spatial    & 22.9\% & 0.780 & 0.825 & 0.644 & 0.705 & 0.668 \\
state-only & 20.5\% & 0.779 & 0.697 & 0.541 & 0.601 & 0.547 \\
full       & 18.9\% & 0.762 & 0.711 & 0.598 & 0.641 & 0.603 \\
\bottomrule
\end{tabular}
\end{table}

\begin{table}[t]
\centering
\caption{\textbf{Overall metric pressure by action.} Rows summarize re-observation-support, visual, and re-observed-consistency metrics by action.}
\label{tab:analysis-action-pressure}
\WRTableAnalysisSetup
\begin{tabular}{lrrrrrr}
\toprule
\WRTableHead
Action & \shortstack{Reobs.\\support} & Integ. & \shortstack{Vis.\\spatial} & \shortstack{Vis.\\state} & \shortstack{Reobs.\\spatial} & \shortstack{Reobs.\\state} \\
\midrule
fold  & 19.1\% & 0.789 & 0.844 & 0.652 & 0.677 & 0.599 \\
jump  & 18.0\% & 0.657 & 0.789 & 0.637 & 0.726 & 0.649 \\
knock & 24.2\% & 0.643 & 0.812 & 0.624 & 0.656 & 0.568 \\
place & 21.9\% & 0.810 & 0.791 & 0.605 & 0.673 & 0.627 \\
sit   & 21.0\% & 0.819 & 0.747 & 0.574 & 0.642 & 0.604 \\
tip   & 24.1\% & 0.706 & 0.787 & 0.575 & 0.717 & 0.640 \\
\bottomrule
\end{tabular}
\end{table}

\begin{figure}[t]
\centering
\includegraphics[width=\linewidth]{\detokenize{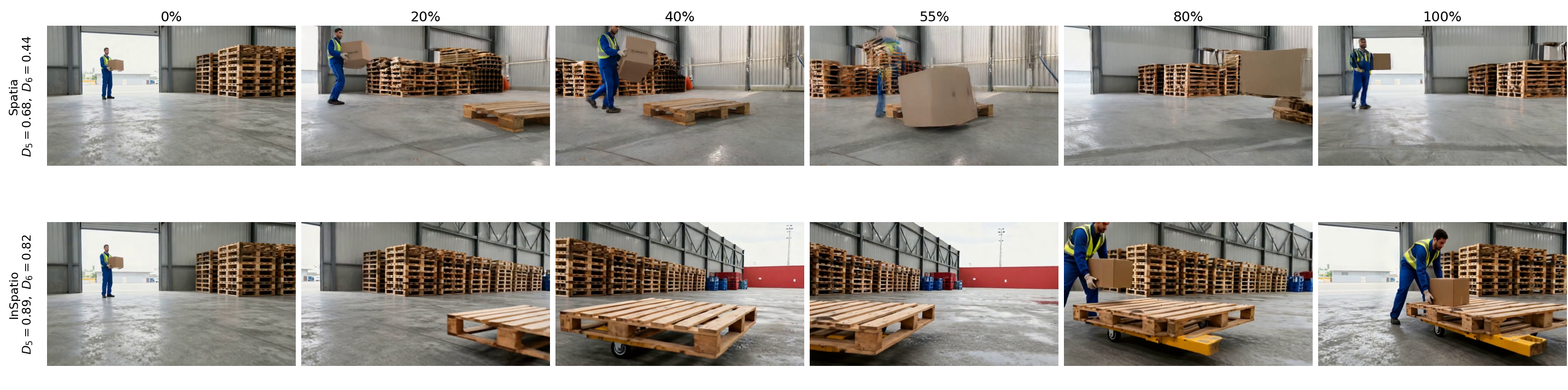}}
\caption{\textbf{Re-observed consistency under matched re-observation (loading dock).} A loading-dock box-placement task (full event, R$\rightarrow$L yaw) on two source-video interfaces, where both rows create judgeable re-observation. \emph{Top (Spatia):} the box returns mislocated and malformed, so re-observed consistency stays low ($D_5=0.68$, $D_6=0.44$). \emph{Bottom (InSpatio):} the box returns correctly placed on the pallet, with high re-observed consistency ($D_5=0.89$, $D_6=0.82$). Judgeable re-observation alone does not guarantee a correct return.}
\label{fig:analysis-endpoint-low-high-case}
\end{figure}

\begin{figure}[t]
\centering
\includegraphics[width=\linewidth]{\detokenize{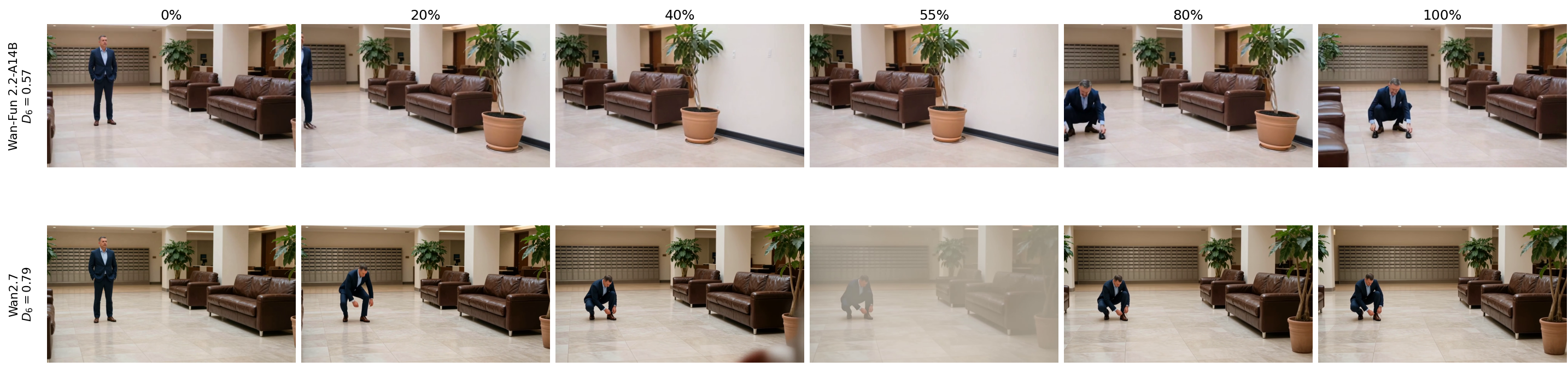}}
\caption{\textbf{State-only re-observed contrast (Wan-Fun 2.2-A14B vs.\ Wan2.7).} A fixed sofa-sitting task (state-only event, R$\rightarrow$L yaw) under matched, judgeable re-observation. \emph{Top (Wan-Fun 2.2-A14B):} the subject returns crouched on the open floor away from the sofa, so re-observed-state consistency stays low ($D_6=0.57$). \emph{Bottom (Wan2.7):} the subject returns seated at the sofa with higher re-observed-state consistency ($D_6=0.79$). State-only pressure therefore persists inside re-observed-state consistency even after re-observation is available.}
\label{fig:analysis-stateonly-case}
\end{figure}

\begin{figure}[t]
\centering
\includegraphics[width=\linewidth]{\detokenize{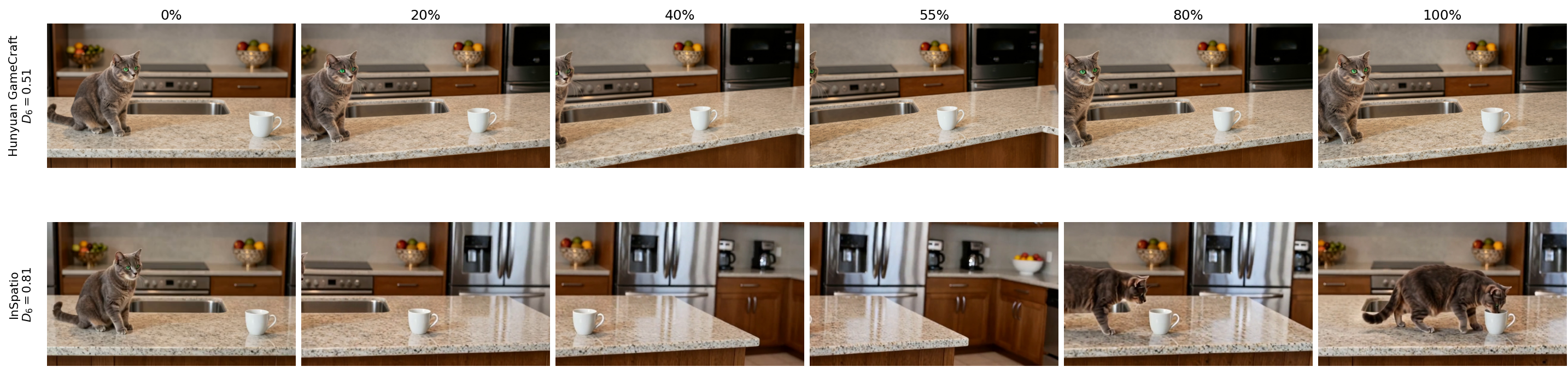}}
\caption{\textbf{Re-observed event-state contrast (kitchen cup knock).} The same cup-knock task (spatial event, R$\rightarrow$L yaw) on two interfaces. \emph{Top (Hunyuan GameCraft):} the knocked cup returns in an inconsistent state, so re-observed event-state stays low ($D_6=0.51$). \emph{Bottom (InSpatio):} the cup's re-observed state is reconstructed more consistently ($D_6=0.81$). Re-observed event-state consistency varies sharply across models under a fixed event and yaw.}
\label{fig:analysis-knock-case}
\end{figure}

Selected family-by-action slices were used only for case mining. The compact appendix keeps the action-level pressure table and the per-sample montages in Figures~\ref{fig:analysis-direction-support-case} through~\ref{fig:analysis-knock-case}, rather than adding another slice table.

\section{Frame-Level Failure Forensics}
\label{app:forensics}

The findings in Sections~\ref{sec:metric_dataset_common} and~\ref{sec:model_subtype} are summary statistics over the $9{,}600$ evaluated clips. This appendix grounds every finding in the raw video. For each claim we fix a scene and vary exactly one factor, then read evenly spaced frames from the generated clip to show \emph{what the picture does} where the metric drops. All frames are decoded directly from the released per-sample outputs (the row manifest in Appendix~\ref{app:repro}); the absolute clip paths are recorded as comments next to each figure for reproducibility, and the scores in the row labels are the same per-clip $D_2$--$D_6$ values used in the aggregate analysis. We cover, in order: visual quality versus a correct return (Section~\ref{sec:metric_dataset_common}, Finding~1); the in-place-change penalty and its distinct failure modes (Finding~2); camera-channel access (Section~\ref{sec:model_subtype}); and the shared in-place failure across interfaces (Section~\ref{sec:model_subtype}). Throughout, $D_2$ is visual integrity, $D_3$/$D_4$ are visible spatial/state, and $D_5$/$D_6$ are re-observed spatial/state.

\subsection{Visual quality does not certify a correct return}
\label{app:forensic-decouple}

\begin{figure}[t]
\centering
\includegraphics[width=\linewidth]{\detokenize{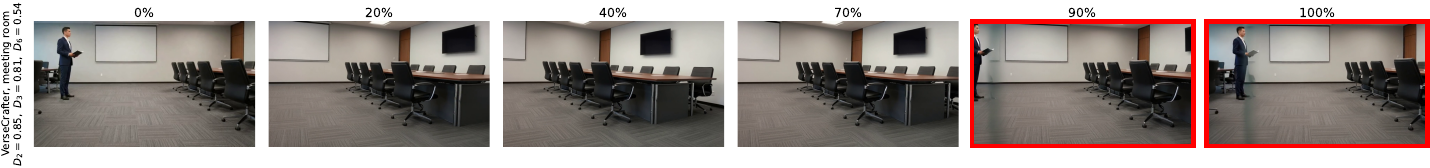}}
\caption{\textbf{Visual integrity is not world-state correctness (VerseCrafter).} A meeting-room clip keeps high visual integrity ($D_2=0.85$) and accurate visible layout ($D_3=0.81$) under a correct yaw, yet the prompted \emph{sit} never happens: on return the subject is still standing, so re-observed-state collapses ($D_6=0.54$).}
\label{fig:forensic-decouple}
\end{figure}

Finding~1 (Section~\ref{sec:metric_dataset_common}) claims that a clean image, a correct camera move, and a subject that returns are three separate successes that need not compose. Figure~\ref{fig:forensic-decouple} shows a single clip in which all three hold yet the return is wrong. VerseCrafter renders the meeting-room scene at high visual integrity ($D_2=0.85$) and executes the requested yaw with an accurate visible layout of the conference table and chairs ($D_3=0.81$), and the person re-enters the frame at the end of the sweep. But the return is wrong: the prompted event was to \emph{sit on a chair}, yet when the subject comes back into view the person is still standing by the screen rather than seated, so the returned world state no longer matches the event that should have occurred while the subject was out of view, and re-observed-state collapses to $D_6=0.54$. High image quality and a correct camera path are therefore certificates of an in-frame continuation, not of a correct reappearance; only the re-observed-consistency probes catch the difference. This is the per-clip face of the loosely-coupled metric structure in Figure~\ref{fig:analysis-corr-heatmap}.

\subsection{Why an in-place change degrades across re-observation}
\label{app:forensic-inplace}

\begin{figure}[t]
\centering
\includegraphics[width=\linewidth]{\detokenize{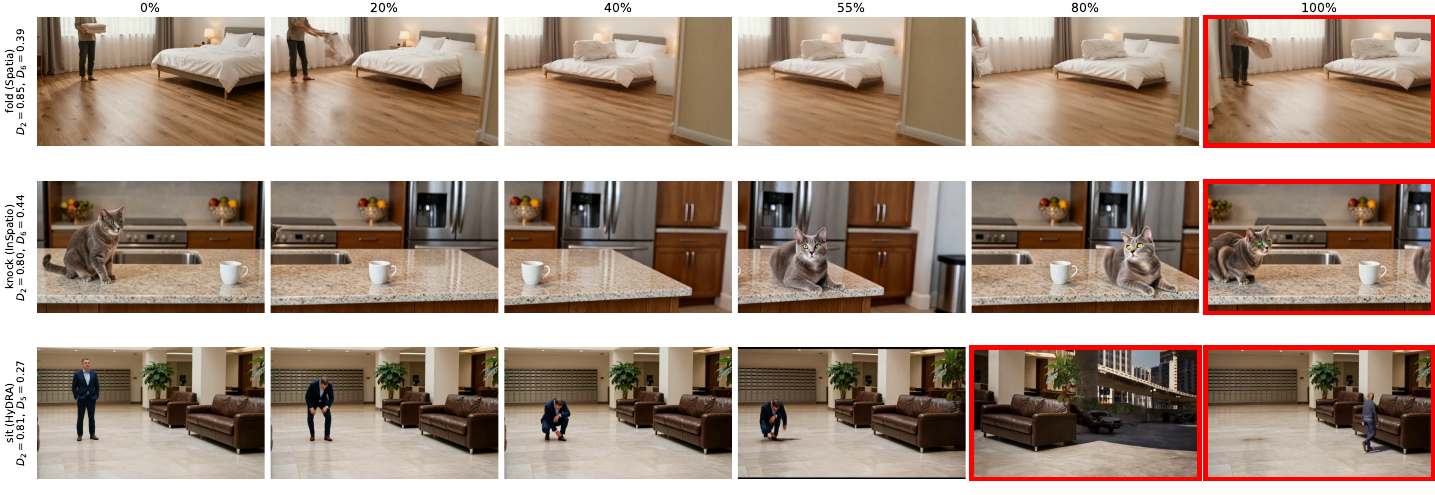}}
\caption{\textbf{Distinct in-place failure modes across actions.} Each row is one in-place clip on a strong model under a right-to-left yaw, with the red frame marking the failure: \emph{fold} returns a wrong fold state, \emph{knock} returns the cup \emph{upright} so the change is silently erased, and \emph{sit} loses the seated person during occlusion ($D_5=0.27$). Different surface artifacts share one cause: no anchor for an unobserved change.}
\label{fig:forensic-modes}
\end{figure}

The in-place penalty is not a single artifact but a family of failures that share one cause, because there is no new coordinate to anchor the unobserved change. Figure~\ref{fig:forensic-modes} samples three in-place actions across strong models and finds three distinct signatures. In \emph{fold} (Spatia), the person returns but the blanket comes back in an unresolved, wrong fold state. In \emph{knock} (InSpatio), the most telling mode appears: the cup returns \emph{upright}, as if the event never happened, because the carrier replays the last observed intact configuration and silently erases the change. In \emph{sit} (HyDRA), the failure is loss of the target: the seated person dissolves during occlusion and the scene briefly collapses, then a person reappears displaced, driving re-observed-spatial to $D_5=0.27$. Wrong-state, erasure, and target-loss are different surface forms of the same missing capability: binding an endpoint that was never in view.

\subsection{Why access depends on the camera channel}
\label{app:forensic-access}

\begin{figure}[t]
\centering
\includegraphics[width=\linewidth]{\detokenize{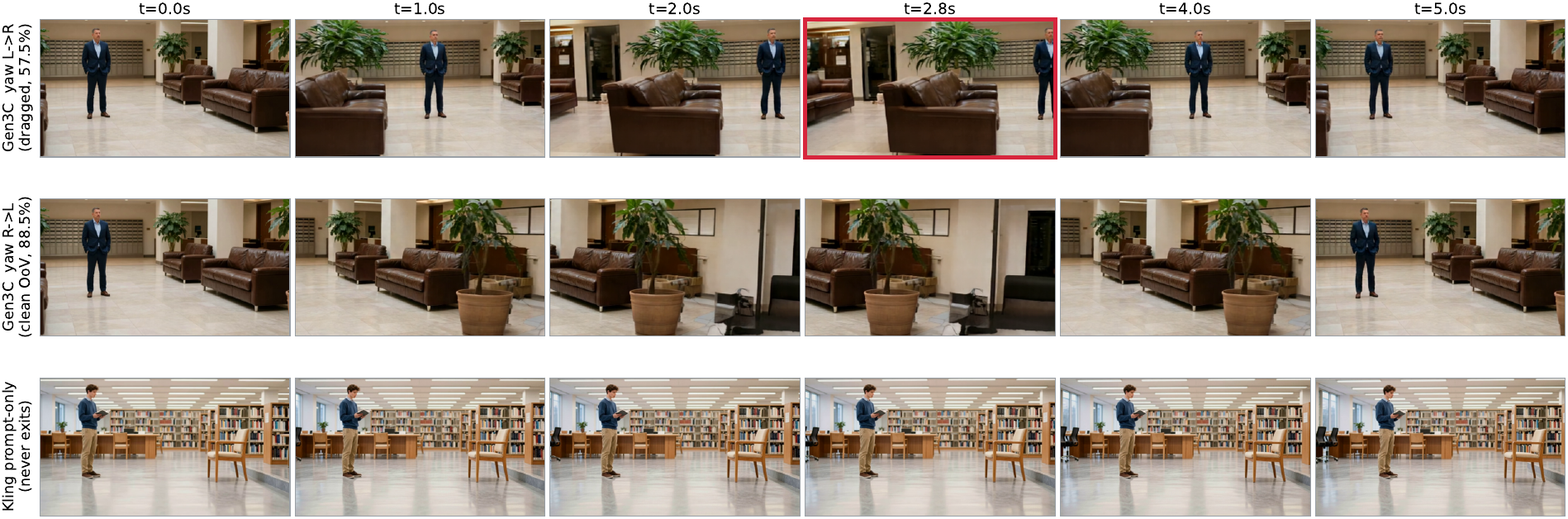}}
\caption{\textbf{Re-observation access is set by the camera channel.} Evenly spaced frames over one camera sweep. \emph{Top (Gen3C, L$\rightarrow$R):} the camera does not swing far enough to move the subject out of frame, so no hidden-then-returned event is created ($57.5\%$ support). \emph{Middle (Gen3C, R$\rightarrow$L):} the same scene cleanly hides and returns the subject ($88.5\%$ support), the $0.65$ retention of Table~\ref{tab:yaw_retention}. \emph{Bottom (Kling, prompt-only):} a clean clip ($D_2\approx0.93$) whose timid camera keeps the subject framed, so no hidden-then-returned event is posed.}
\label{fig:forensic-access}
\end{figure}

Figure~\ref{fig:forensic-access} grounds the access half of the analysis: whether a clip ever creates a hidden-then-returned event at all. The top two rows fix a Gen3C lobby scene and flip only the yaw direction. Right-to-left (middle row) sweeps the standing person cleanly out of frame, so that the empty lobby is visible through the occlusion, and brings the person back, producing a valid re-observation; this direction yields re-observation support of $88.5\%$. Left-to-right (top row) does not: in this direction the prescribed yaw does not sweep the subject out of frame, so the target stays visible throughout and no hidden-then-returned event is created. This is a layout effect rather than a model failure---the subject's starting position clears the view only under the opposite yaw---which is why Gen3C's $0.65$ retention ratio (Table~\ref{tab:yaw_retention}) serves as the geometry-calibrated baseline for reading other models' directional gaps.

The bottom row shows the opposite extreme. A prompt-only system (Kling) renders a clean, high-integrity clip ($D_2\approx0.93$) but executes such a timid camera motion that the reading student never leaves the frame across the entire sweep; no occlusion, no return, and therefore no judgeable re-observation. This is why prompt-only interfaces top the visible scores yet sit at the bottom of re-observation support (Section~\ref{sec:model_subtype}, Table~\ref{tab:gen3c-baseline}): their bottleneck is not rendering quality but the failure to create the test. Together the three rows make the central access claim concrete: camera motion governs whether the hidden-state question can be asked, and that is decided by the camera channel, not by visible image quality.

\subsection{All interfaces fail on the same in-place case}
\label{app:forensic-crossinterface}

\begin{figure}[t]
\centering
\includegraphics[width=\linewidth]{\detokenize{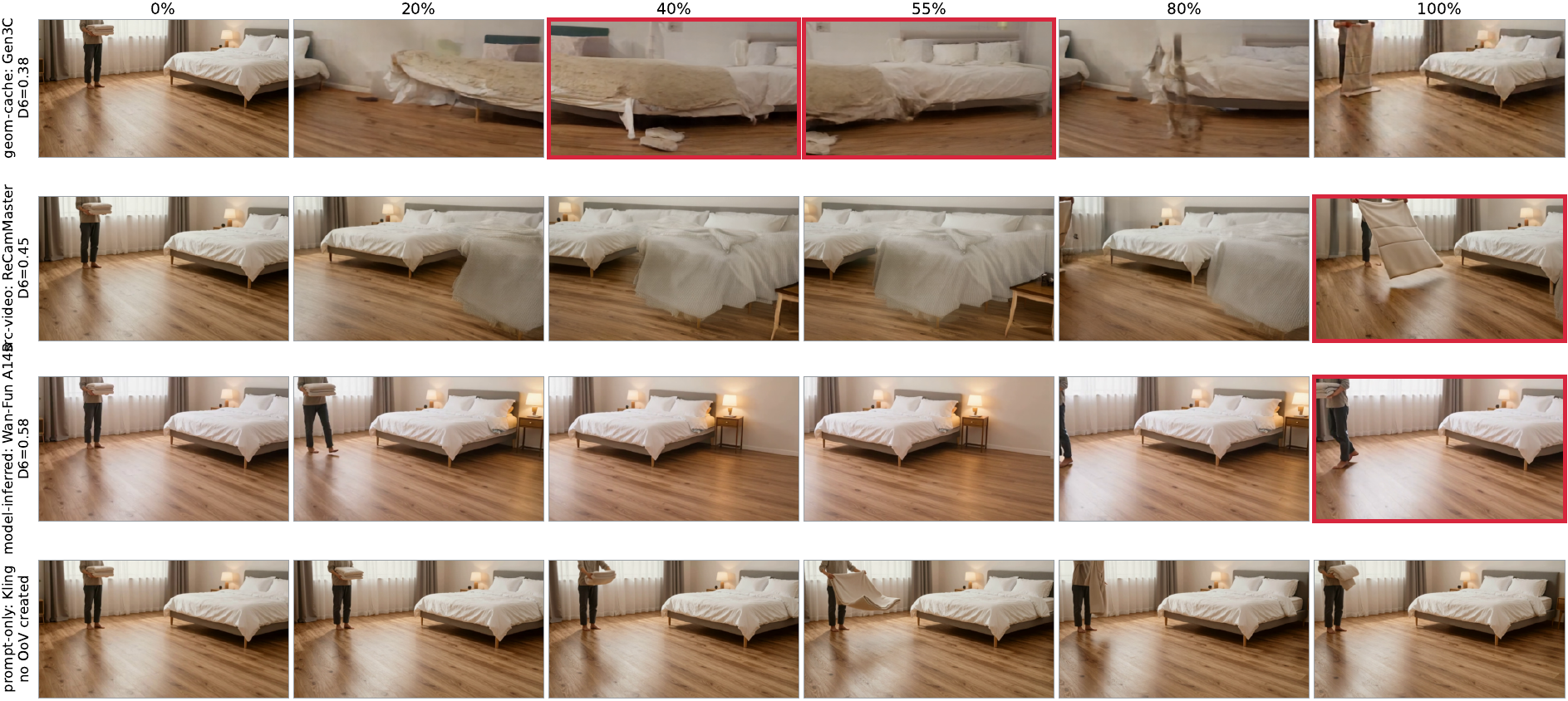}}
\caption{\textbf{One in-place event, four interfaces, four failure signatures.} The same blanket-fold event and camera path run through each interface type. \emph{Geom-cache (Gen3C):} the bed and blanket melt into warped geometry (red frames, $D_6=0.38$). \emph{Source-video (ReCamMaster):} as the camera sweeps out, two separate beds are hallucinated into one merged structure ($D_6=0.45$). \emph{Model-inferred (Wan-Fun A14B):} cleanest frame ($D_2=0.89$) but the subject is dragged along the camera direction ($D_6=0.58$). \emph{Prompt-only (Kling):} the subject never exits, so no re-observation is created. No interface recovers the unobserved in-place change.}
\label{fig:forensic-crossinterface}
\end{figure}

The access advantage of source-video and geometry-cache interfaces (Section~\ref{sec:model_subtype}) buys the right to be tested; it does not buy a correct return. Figure~\ref{fig:forensic-crossinterface} fixes one in-place event, folding a blanket on a bed, and one camera path, then runs it through all four interface types. None recovers the folded state, and each fails in the signature of its own mechanism. The geometry-cache interface (Gen3C) melts the bed and blanket into warped geometry as it re-projects a cache that never observed the fold ($D_6=0.38$). The source-video interface (ReCamMaster) hallucinates as the camera sweeps out, fusing two separate beds into one merged structure instead of preserving the single bed ($D_6=0.45$). The model-inferred interface (Wan-Fun A14B) keeps the cleanest frame of the four ($D_2=0.89$) but couples the subject to ego-motion, dragging the person along the camera direction instead of letting the fold be re-observed ($D_6=0.58$). The prompt-only interface (Kling) never lets the subject leave the frame, so it produces no re-observation to score at all. Across four different mechanisms, namely geometric melt, hallucinated structure, camera--subject coupling, and no test created, no interface returns the correctly folded state, so the unobserved in-place endpoint is preserved by none of them.

\subsection{High Scores, Frame-Level Failures: The Metric Paradox}
\label{app:forensic-highscore}

\begin{figure}[t]
\centering
\includegraphics[width=\linewidth]{\detokenize{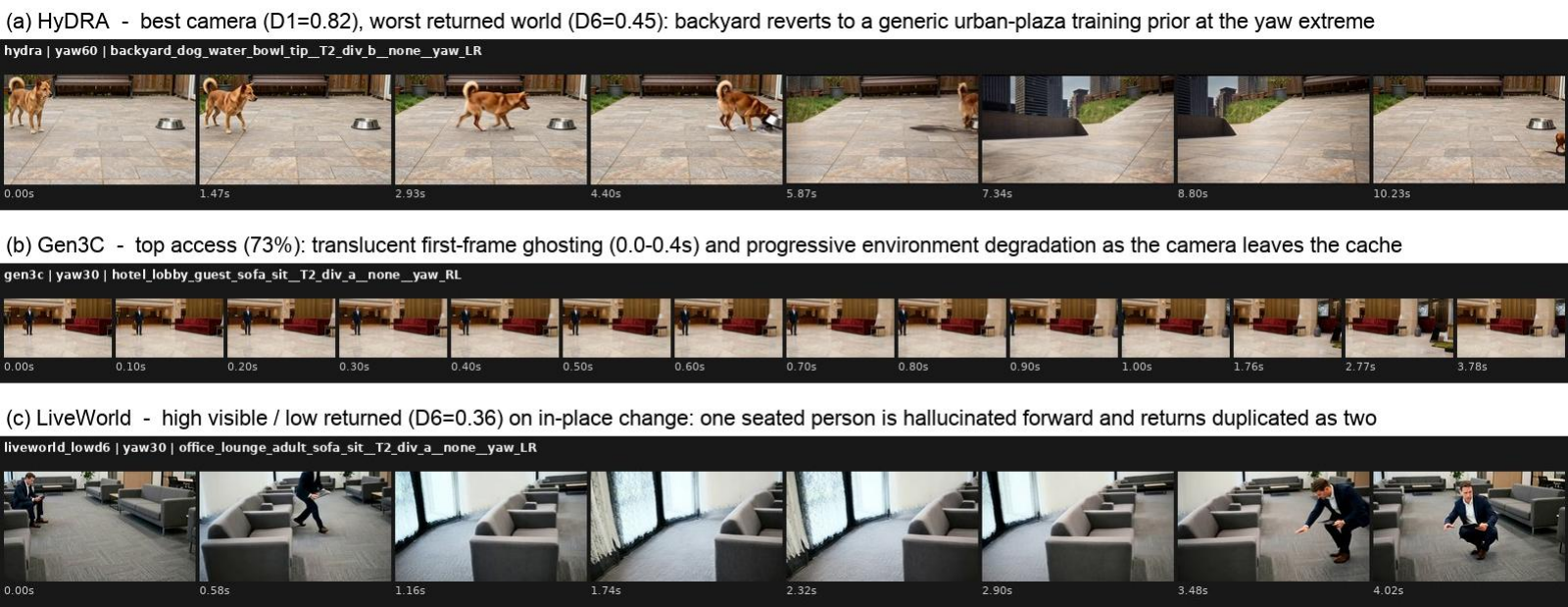}}
\caption{\textbf{High scores, frame-level failures.} Each row is a high scorer on one axis whose failure lives on another. \emph{(a) HyDRA} ($D_1=0.822$, best; $D_6=0.445$, worst): at the yaw extreme the real backyard is replaced by a generic urban plaza as the synthetic-domain training prior takes over the unobserved region. \emph{(b) Gen3C} (access $73\%$, best): translucent first-frame ghosting ($0.0$--$0.4$\,s) and progressive degradation once the camera leaves the cached frustum. \emph{(c) LiveWorld} (visible spatial ${>}0.9$ on easy clips; $D_6{\approx}0.31$--$0.40$ on in-place changes): the out-of-sight subject is hallucinated forward and returns \emph{duplicated}, one seated person re-synthesized as two, while the room stays sharp.}
\label{fig:forensic-highscore}
\end{figure}

Aggregate dimension scores can rank a model at the top of one axis while a different axis quietly collapses; the headline number and the failure then live on separate axes, and only frame-level reading separates them. Three of the strongest single-axis scorers make the point (Figure~\ref{fig:forensic-highscore}).

\paragraph{HyDRA: best camera execution, worst returned world.} HyDRA controls the camera better than any other row yet returns the least faithful world. It posts the highest requested-camera precision of all 23 rows ($D_1=0.822$) and strong common-yaw alignment ($0.855$), yet the weakest re-observed state of any model ($D_6=0.445$) and a low visual integrity ($D_2=0.691$). The frames explain the split (Figure~\ref{fig:forensic-highscore}a). HyDRA's hybrid memory is trained only on the fully synthetic, engine-rendered HM-World corpus, so it executes the prescribed camera trajectory precisely, but once the yaw exposes a region the source never carried, it fills that region from its synthetic training prior rather than the real scene, so that a backyard becomes a generic paved urban plaza with skyscrapers or the view collapses into a dark void. Precise camera control therefore coexists with a hallucinated world: the model is overfit to its training domain exactly where WRBench probes the unobserved.

\paragraph{Gen3C: top access, first-frame ghosting and progressive degradation.} Gen3C leads re-observation support ($73\%$) because its 3D cache re-projects the seen scene. The same reprojection pipeline, however, leaves two frame-level artifacts that the aggregate $D_2$ ($0.749$) only partly reflects (Figure~\ref{fig:forensic-highscore}b). The first ${\sim}0.4$\,s is a translucent, doubled overlay while the cache initializes, and once the camera leaves the cached frustum the scene degrades into smears, darkening, and object ghosting. Access is bought with a cache whose generative infill is not yet photometrically closed.

\paragraph{LiveWorld: a plausible reappearance of the wrong world.} LiveWorld's easy clips return cleanly, so its aggregate visible scores are high (visible spatial ${>}0.9$); the failure concentrates on the in-place-change events, where re-observed state falls to $D_6{\approx}0.31$--$0.40$. There the cost of its monitor-agent design becomes visible, because it rolls the backbone forward to \emph{hallucinate} the out-of-sight subject rather than retrieving a stored state (Figure~\ref{fig:forensic-highscore}c): a single seated person is re-synthesized on return as \emph{two} crouching figures, or an extra displaced body is spawned near the original, while the static room stays sharp. The subject is generated forward, not preserved, so the model returns a confident but duplicated occupant and the event endpoint is lost.

\section{Model Subtype and Series Reading Guide}
\label{app:model-subtype-series}

The analysis is organized around metric--dataset common problems, model subtype and viewpoint condition type, targeted model series, human-aligned interpretation, and future reward or policy-training use. This appendix records the assignment rules used for the condition-type and series comparisons. The generated artifacts use V2V, TI2V, and I2V labels; methods that take text plus a source/reference video are assigned to the source-video condition rather than a separate top-level family.

\subsection{Subtype Taxonomy}
\label{app:model-subtype-taxonomy}

The subtype taxonomy is a claim-scoping device, not a new aggregation script, and Table~\ref{tab:appendix-condition-groups} summarizes the resulting grouping. Its primary organizing variable is the dominant viewpoint condition type: the form of input that provides information about the requested viewpoint change. Source-video rows use a reference stream for appearance, layout, and partial event evidence. Geometry-cache rows use point clouds, 3D caches, or 3D--4D controls to make parts of camera-target access computable. Model-inferred rows have no external view-state reference and must synthesize the new view under local camera, action, or state controls. Prompt-only rows expose natural-language camera intent rather than a certified requested-control interface and therefore use CamAlign/common-yaw diagnostics rather than strict CamPrec. Camera-control interfaces and dynamic-state evidence sources are recorded as modifiers, not as peer subtype names.

The unit of assignment is the evaluated WRBench row, not the model family. If a future system exposes multiple interfaces, it should be assigned by the interface and evidence path used in the reported WRBench run. For hybrids, an external source-video or geometry-cache condition takes precedence when it supplies the view-state information used for re-observation; otherwise local camera/action/state modules are recorded as modifiers within the model-inferred condition. Proprietary API rows remain prompt-only unless strict requested-control artifacts are available.

Representative source-video rows cite ReCamMaster~\citep{bai2025recammaster}, HyDRA~\citep{chen2026out}, and InSpatio World 14B~\citep{team2026inspatio}. Geometry-cache rows cite Gen3C~\citep{ren2025gen3c}, Spatia~\citep{zhao2026spatia}, and VerseCrafter~\citep{zheng2026versecrafter}. Model-inferred rows cite Wan-Fun~\citep{wan2025wan}, LingBot~\citep{team2026advancing}, LiveWorld~\citep{duan2026liveworld}, Hunyuan GameCraft~\citep{li2025hunyuan}, Hunyuan WorldPlay~\citep{sun2025worldplay}, and MagicWorld~\citep{li2025magicworld}; prompt-only product rows without paper entries are left uncited.

\begin{table}[t]
\centering
\caption{\textbf{Appendix viewpoint condition taxonomy.} Rows are grouped by viewpoint condition type; control and state modules are modifiers.}
\label{tab:appendix-condition-groups}
\WRTableSetup
\begin{tabular}{L{0.24\linewidth} L{0.31\linewidth} L{0.37\linewidth}}
\toprule
\WRTableHead
Viewpoint condition type & Representative rows & Assignment rule \\
\midrule
Source-video condition & ReCamMaster, HyDRA, InSpatio World 14B & Externalized appearance/layout evidence. \\
Geometry-cache condition & Gen3C, Spatia, VerseCrafter & Computable access; static-memory channel. \\
Model-inferred condition & Wan-Fun, LingBot, LiveWorld, Hunyuan GameCraft, Hunyuan WorldPlay, MagicWorld & Local generation with modifiers. \\
Prompt-only condition & Hailuo, Kling, Wan API rows, HappyHorse & CamAlign only; no strict CamPrec. \\
\bottomrule
\end{tabular}
\end{table}

The series comparisons are diagnostic contrasts rather than scalar rankings. LingBot versus Wan-Fun separates visible preservation from camera-target access; Wan scaling separates the returned-evidence frontier from endpoint binding; source, memory, geometry, and local-control routes separate access mechanisms; and the Hunyuan split illustrates that interface compatibility is not the same as re-observed consistency.

\begingroup
\WRTableWideSetup
\setlength{\tabcolsep}{2pt}
\renewcommand{\arraystretch}{1.04}
\setlength{\LTleft}{0pt}
\setlength{\LTright}{0pt}
\captionof{table}{\textbf{Wan-derived model increments behind the series diagnostics.}
Rows are grouped by documented Wan base or reported Wan-backed release family; ``Unreported'' marks public-release gaps rather than negative evidence.}
\label{tab:appendix-wan-increment-taxonomy}
\begin{longtable}{@{}L{0.13\textwidth} L{0.185\textwidth} L{0.17\textwidth} L{0.28\textwidth} L{0.185\textwidth}@{}}
\toprule
\WRTableHead
Evaluated model & Architecture & Training update & Data evidence & Diagnostic mechanism \\
\midrule
\endfirsthead
\toprule
\WRTableHead
Evaluated model & Architecture & Training update & Data evidence & Diagnostic mechanism \\
\midrule
\endhead
\bottomrule
\endfoot
\bottomrule
\endlastfoot
\WRSubGroup{5}{Wan2.1-T2V-1.3B} \\
Wan-Fun 2.1-1.3B & Control-Camera camera-lens conditioning & Public fine-tuning scope unreported & Camera-control data unreported & Camera condition only; no public camera/content disentanglement or state supervision \\
ReCamMaster & Source-video rerendering plus target-camera encoder & Camera encoder and 3D-attention layers trained; base mostly frozen & MultiCamVideo: 13,600 scenes, 10 synchronized cameras, paired-view sampling; toy$\to$full ablation FVD $179\to123$ & Same event under multiple cameras separates camera motion from scene content; paired-view diversity is the critical training factor \\
HyDRA & Memory Tokenizer plus Dynamic Retrieval Attention & Memory and retrieval modules; 10,000 iterations; affinity-vs-FOV retrieval ablation $+0.018$ subject consistency & HM-World: 59,000 clips, fully engine-rendered (17 scenes, 49 subjects), decoupled camera/subject trajectories and exit-entry events & Reappearance memory under camera/subject decoupling; synthetic-domain prior fills the unobserved region (App.~\ref{app:forensic-highscore}); endpoint writing is separate \\
\WRSubGroup{5}{Wan2.1-T2V-14B} \\
Wan-Fun 2.1-14B & Control-Camera camera-lens conditioning & Public fine-tuning scope unreported & Camera-control data unreported & Same camera condition at larger Wan2.1 scale; no new endpoint supervision \\
VerseCrafter & GeoAdapter with rendered 4D control maps & GeoAdapter only; Wan encoder, diffusion transformer, and decoder frozen; 3D-Gaussian-traj ablation beats bbox/point (ObjMC $2.51$ vs $4.52$/$6.90$) & VerseControl4D: 35,000 train / 1,000 validation; rendered RGB/depth, trajectories, masks & External 4D geometry supplies camera/object motion without backbone retraining; trajectories are prescribed, not learned out-of-sight state \\
LiveWorld & Monitor-agent generative out-of-sight evolution (VACE state adapter + LoRA on frozen Wan-14B); not test-time optimization & State adapter 10,000 steps; rank-64 low-rank adapter 5,000 steps; w/o-Evo ablation VQA $59\to18$ & MIRA, RealEstate10K, and SpatiaVID\_HQ examples; full mix partially public & Out-of-sight dynamics hallucinated forward by monitor rollouts; returns a duplicated subject on in-place change (App.~\ref{app:forensic-highscore}) \\
\WRSubGroup{5}{Wan2.1-I2V-14B-480P} \\
InSpatio World 14B & STAR video-to-video control with DA3 depth and point-cloud rendering & JDMD dual-teacher distillation (synthetic V2V + real T2V; no RL stage); 14B recipe unreported & Reference video, DA3 depth, point-cloud rendering, trajectory text; official config names the I2V-14B generator & Geometry/source-video signal separates view access from internal state writing \\
\WRSubGroup{5}{Wan2.2-TI2V-5B} \\
Wan-Fun 2.2-5B & Control-Camera conditioning on dense TI2V-5B & Public fine-tuning scope unreported & Camera-control data unreported; Wan2.2 reports +65.6\% images, +83.2\% videos, aesthetic filtering & Camera condition plus stronger visible prior; control and base-data effects coupled \\
Spatia & Persistent 3D point-cloud memory with VACE controls & Control blocks about 8,000 steps; rank-64 low-rank adapter about 5,000 steps & RealEstate 40,000 plus SpatialVID-HD 10,000; pose estimation, dynamic-object removal & Spatial memory stores layout and camera path, not dynamic event endpoints; SAM2 strips dynamics before the SLAM update, so the cloud is static-only \\
\WRSubGroup{5}{Wan2.2-I2V-A14B/MoE family} \\
Wan-Fun 2.2-A14B & Control-Camera conditioning on Wan2.2-Fun-A14B & Public fine-tuning scope unreported & Camera-control data unreported; HF metadata/config identify I2V-A14B high/low-noise MoE lineage & Camera condition under I2V-A14B/MoE prior; control data remain unreported \\
LingBot World & Base-Cam pose-conditioned DiT modulation & Family-level pre-training prior; middle training for memory/action; causal and few-step post-training & Family-level data evidence: real/open videos, game logs, Unreal synthetic trajectories, synchronized camera/action data, captions; Base-Cam split unreported & Explicit per-frame pose competes with subject-continuation prior; memory is emergent from long context, with no explicit out-of-sight module \\
LingBot Act & Base-Act action adapters with continuous camera & Family-level action adapters and adaptive layer normalization; main diffusion-transformer blocks frozen & Family-level data evidence: hybrid game/simulation action data central; act-preview split unreported & Action-conditioned state channel, distinct from camera-pose conditioning \\
\end{longtable}
\begin{minipage}{\textwidth}
\raggedright
\WRTableNote{Base groups use the authors' release-family labels when documented: T2V, I2V, TI2V, A14B, and MoE are retained only as compact base identifiers. ReCamMaster's evaluated open port is Wan2.1-T2V-1.3B; its paper model used an internal text-to-video backbone. InSpatio's official inference config names a Wan2.1-I2V-14B-480P generator for the 14B V2V path, while the released training recipe remains unreported. Wan-Fun 2.2-A14B is grouped with I2V-A14B/MoE because HF metadata/config identify Wan2.2-I2V-A14B high/low-noise MoE lineage, while public Fun documentation reports the release as A14B Control-Camera and leaves fine-tuning scope unreported. LingBot rows use I2V-A14B/MoE inference, but their training/data entries are family-level evidence rather than checkpoint-specific recipes. Runtime profiles such as frame count and frame rate are omitted because they describe the generation interface, not the model-side increment.}
\end{minipage}
\endgroup

\subsection{Series Comparison Notes}
\label{app:series-comparison-notes}

\paragraph{LingBot vs Wan-Fun.}
LingBot~\citep{team2026advancing} and Wan-Fun~\citep{wan2025wan} isolate the visible-preservation versus camera-target-displacement tradeoff inside local TI2V control. LingBot World and LingBot Act have the strongest local visible spatial/state scores, but their re-observation support is only $6.0\%$/$6.4\%$, so their re-observed-consistency values are sparse-re-observation readings. Wan-Fun executes stronger camera displacement and reaches higher re-observation support, but its visible event scores are lower than LingBot's. This comparison isolates how camera motion couples to target displacement across access routes.

\paragraph{Wan scaling diagnostic.}
Wan-Fun scaling provides a within-family diagnostic rather than a universal capacity rule. In both Wan 2.1 and Wan 2.2, the larger/backbone-upgraded variant improves or preserves visible rendering and raises re-observation support more clearly than it improves conditional re-observed consistency. The diagnostic reading is specific: larger backbones can expand the set of judgeable re-observed samples without automatically binding the hidden event endpoint.

\paragraph{Architectural state-carrier ladder (full roster).}
Coding each evaluated increment by its \emph{state carrier}, the module that decides whether an out-of-view region can be revisited, rather than by its camera-encoding format, orders the full roster as one ladder (Table~\ref{tab:appendix-wan-increment-taxonomy}). Camera-lens control (Wan-Fun 2.1-1.3B/14B, 2.2-5B/A14B) aims the view but stores nothing out of sight (re-observation support $12.0$--$18.2\%$; re-observed state $0.62$--$0.66$). External geometry adapters add a static spatial record, namely VerseCrafter's GeoAdapter over rendered 4D maps ($28.0\%$, re-observed state $0.584$) and Spatia's persistent point cloud ($25.8\%$, $0.586$); these sit among the lowest re-observed state in the series, above only HyDRA. Source-video carriers stream appearance and dynamics for the highest access (ReCamMaster $58.5\%$/$0.616$; InSpatio $62.3\%$/$0.664$) but borrow temporal evidence from the condition. Explicit reappearance memory (HyDRA's Memory Tokenizer and Dynamic Retrieval Attention) earns the highest requested-camera precision ($0.822$) and the weakest re-observed state ($0.445$). A dedicated out-of-sight state adapter (LiveWorld) reaches mid access ($39.6\%$) without closing the endpoint ($0.600$). Pose- and action-conditioned rows (LingBot World/Act) preserve the strongest visible spatial/state consistency among the Wan-derived rows ($0.876/0.735$ and $0.874/0.719$) while camera execution collapses (requested-camera precision $0.513/0.468$; support $6.0\%$/$6.4\%$; returned rows $n{=}30/32$). The ladder is monotonic in access repair but flat in endpoint binding: every carrier stores or replays where a scene can be revisited, none yet stores what resolved there while hidden, and the camera-encoding format does not predict re-observed state.

\paragraph{Public training-signal ladder (full roster).}
The same rows order by the public training signal they expose (Figure~\ref{fig:series-diagnostics}c); unreported recipes are gaps, not negative evidence. Wan-Fun control-camera fine-tuning (scope and data unreported) supervises camera execution only. ReCamMaster's MultiCamVideo ($13{,}600$ scenes, ten synchronized cameras, paired-view sampling) supervises camera/content disentanglement but is tied to that paired distribution. Rendered-geometry training supervises external layout: VerseCrafter trains GeoAdapter alone with the backbone frozen (VerseControl4D, $35{,}000$ clips), and Spatia trains control blocks and a low-rank adapter over RealEstate and SpatialVID-HD, while InSpatio's depth/point-cloud source path keeps its 14B recipe unreported. HyDRA trains memory and retrieval on HM-World ($59{,}000$ clips with exit--entry events) for reappearance; LiveWorld trains a VACE-initialized state adapter for an out-of-sight channel. LingBot's long-trajectory world-model curriculum most directly supervises objects evolving and reappearing within a sequence, which is why it holds visible world state best, though its re-observed metrics are sparse-re-observation readings ($n{=}30/32$). No evaluated row exposes public supervision on whether a contact, containment, posture, or collision outcome survives non-observation; that zero-count tier, together with the ReCamMaster-versus-LingBot contrast, motivates the long-to-short hypothesis and the endpoint-directed reward/policy outlook of Appendix~\ref{app:preference-pair-export}.

\paragraph{Test-time and post-training paradigm (axes beyond camera encoding).}
The roster also separates on two axes the camera-encoding view misses. First, test-time compute: all rows are amortized feed-forward except LiveWorld, which adds runtime multi-agent orchestration, where VLM-registered monitors roll the backbone forward to \emph{hallucinate} out-of-sight dynamics. This is a distinct paradigm from amortized latent/retrieval memory, and \emph{not} per-instance test-time optimization. Second, post-training: every increment that post-trains does so by distribution-matching distillation (ReCamMaster/InSpatio dual-teacher JDMD, Hunyuan WorldPlay Context-Forcing, LingBot causal-plus-DMD) or reward-weighted DMD (MagicWorld), \emph{not} reinforcement learning; ``reward'' in this literature shapes a distillation batch, not an environment-return objective, and none of it supervises the hidden event endpoint. Third, training-data domain: synthetic paired corpora (ReCamMaster's MultiCamVideo, HyDRA's HM-World) buy strong access and camera-alignment priors but risk a synthetic-domain fallback in unobserved regions (Appendix~\ref{app:forensic-highscore}), whereas real-monocular training (Spatia, VerseCrafter, MagicWorld) trades that for weaker geometric replay. These three axes reframe the series result: the endpoint gap is an unwritten training objective, not a property of the camera interface.

\paragraph{Hunyuan GameCraft vs Hunyuan WorldPlay.}
Hunyuan GameCraft~\citep{li2025hunyuan} and Hunyuan WorldPlay~\citep{sun2025worldplay} function as local-control modifiers. GameCraft's player-action interface is partly mismatched to WRBench's prescribed yaw and target-region re-observation protocol, while WorldPlay's chunk memory is more compatible with re-observation pressure. WorldPlay therefore improves access relative to GameCraft, but it still does not establish re-observed consistency because scene-context memory is not the same as writing the exact contact, containment, posture, or collision endpoint into state.

\subsection{Condition-Type Detail Notes}
\label{app:condition-detail-notes}

This appendix provides the expanded condition-interface analysis behind Section~\ref{sec:model_subtype}. The main text uses the compact evidence chain; the notes below specify what each viewpoint condition type externalizes, what advantage it buys, and why that advantage does not yet amount to hidden endpoint binding.

\paragraph{Source-video condition.}
Source-video rows externalize some world evidence into an input stream. ReCamMaster~\citep{bai2025recammaster}, HyDRA~\citep{chen2026out}, and InSpatio World 14B~\citep{team2026inspatio} instantiate source-world transformation under a target camera path, rather than standalone prompt-only simulation. ReCamMaster and HyDRA supply a source or reference stream that already contains appearance, layout, and temporal evidence, giving a route to re-observation support that does not require an explicit point-cloud cache. The tradeoff is that dynamics are borrowed from the conditioning stream rather than inferred as a persistent hidden state. High re-observation support in this group means that the source/reference condition can make the target region visible again; returned visibility still must be checked separately for spatial and state consistency.

\paragraph{Geometry-cache condition.}
Point clouds, 3D caches, spatial memory, and 4D controls make camera-target access more computable. Gen3C~\citep{ren2025gen3c} is the access reference because its point-cloud/3D-cache condition creates the strongest re-observation support, but its re-observed-consistency values are still not saturated. A previously observed region can be re-rendered, warped, or constrained by a stored spatial representation, which explains why high-re-observation-support rows are dominated by geometry or source-scene conditions. The same channel clarifies the measured variable: an event endpoint that occurs while the relevant evidence is not observed still needs re-observed-state evidence. Spatia~\citep{zhao2026spatia} and VerseCrafter~\citep{zheng2026versecrafter} show the same pattern from another angle: geometry can remember where physical surfaces and viewpoints are, while the abstract endpoint state may remain unbound.

\paragraph{Model-inferred condition.}
Wan-Fun~\citep{wan2025wan}, LingBot~\citep{team2026advancing}, LiveWorld~\citep{duan2026liveworld}, Hunyuan GameCraft~\citep{li2025hunyuan}, Hunyuan WorldPlay~\citep{sun2025worldplay}, and MagicWorld~\citep{li2025magicworld} are grouped together because their row-level modifiers do not create a separate viewpoint condition type. Camera-lens control, pose conditioning, action tokens, chunk memory, state adapters, and history retrieval change the metric signature inside local control. Local camera-control TI2V rows synthesize the scene from a first frame and camera/action controls rather than re-rendering a source stream or a cache. Their visible evidence can be strong, and when a returned target is judgeable their conditional re-observed consistency is not necessarily the weakest signal. The recurring bottleneck is retrievable re-observation support: visible generation quality and returned-evidence availability are supplied by different mechanisms.

\paragraph{Prompt-only condition.}
Prompt-only rows are outside strict requested-control CamPrec. They preserve visible content well and pass static-hold sanity checks, but the common-yaw CamAlign diagnostic is weak and re-observed-consistency rows are often sparse. These rows show that visually plausible prompt-camera continuation can fail the camera-intent bridge. Hailuo and Kling, for example, can keep high visible spatial/state scores while rarely creating the hidden-then-returned evidence needed to evaluate re-observed consistency; the active bottleneck is sparse returned-evidence creation rather than conditional re-observed consistency alone.

\subsection{Directional Access Diagnostic}
\label{app:directional-access-notes}

Yaw-direction slices require a geometry-anchored layout baseline. Gen3C shows that the benchmark layout itself can make one yaw direction easier than the other, so a raw left--right gap is not automatically a model failure. The stronger signal is when a method falls far below the geometry-calibrated retention pattern despite executing camera-like motion. The row-level diagnostic therefore supports condition-type analysis and hard-case mining, while the main text keeps the central claim on access, re-observation support, and conditional re-observed consistency.

\begin{table}[t]
\centering
\WRTableSetup
\caption{\textbf{Geometry-calibrated yaw access.} Entries report re-observation support count over yaw-direction rows, followed by the percentage; retention is L$\rightarrow$R divided by R$\rightarrow$L.}
\label{tab:yaw_retention}
\begin{tabular}{lccc}
\toprule
\WRTableHead
Model & R$\rightarrow$L reobs. & L$\rightarrow$R reobs. & Retention \\
\midrule
Gen3C & 177/200 (88.5\%) & 115/200 (57.5\%) & 0.65 \\
Hailuo 2.3 & 15/100 (15.0\%) & 4/100 (4.0\%) & 0.27 \\
HappyHorse 1.0 I2V & 6/100 (6.0\%) & 25/100 (25.0\%) & 4.17 \\
Hunyuan GameCraft & 15/200 (7.5\%) & 9/200 (4.5\%) & 0.60 \\
Hunyuan WorldPlay & 102/200 (51.0\%) & 0/200 (0.0\%) & 0.00 \\
HyDRA & 95/200 (47.5\%) & 38/200 (19.0\%) & 0.40 \\
InSpatio World 14B & 166/200 (83.0\%) & 83/200 (41.5\%) & 0.50 \\
Kling v2.6 & 10/100 (10.0\%) & 0/100 (0.0\%) & 0.00 \\
Lingbot World Cam & 30/200 (15.0\%) & 0/200 (0.0\%) & 0.00 \\
Lingbot World Act & 32/200 (16.0\%) & 0/200 (0.0\%) & 0.00 \\
LiveWorld & 85/200 (42.5\%) & 113/200 (56.5\%) & 1.33 \\
MagicWorld & 62/200 (31.0\%) & 13/200 (6.5\%) & 0.21 \\
ReCamMaster & 168/200 (84.0\%) & 66/200 (33.0\%) & 0.39 \\
Spatia & 122/200 (61.0\%) & 7/200 (3.5\%) & 0.06 \\
VerseCrafter & 139/200 (69.5\%) & 1/200 (0.5\%) & 0.01 \\
Wan-Fun 2.1-14B & 91/200 (45.5\%) & 0/200 (0.0\%) & 0.00 \\
Wan-Fun 2.1-1.3B & 66/200 (33.0\%) & 3/200 (1.5\%) & 0.05 \\
Wan-Fun 2.2-5B & 57/200 (28.5\%) & 3/200 (1.5\%) & 0.05 \\
Wan-Fun 2.2-A14B & 86/200 (43.0\%) & 2/200 (1.0\%) & 0.02 \\
Wan2.2 I2V Plus & 5/100 (5.0\%) & 1/100 (1.0\%) & 0.20 \\
Wan2.6 I2V & 26/100 (26.0\%) & 12/100 (12.0\%) & 0.46 \\
Wan2.7 I2V & 13/100 (13.0\%) & 5/100 (5.0\%) & 0.38 \\
WanX2.1 I2V Turbo & 1/100 (1.0\%) & 2/100 (2.0\%) & 2.00 \\
\bottomrule
\end{tabular}
\end{table}

\section{Human Validation Notes}
\label{app:human-alignment-notes}

This appendix gives the calibration evidence behind Section~\ref{sec:human_validation}. Human comparison families follow the same reported dimensions used by WRBench: requested-camera precision, prompt-camera alignment, visual integrity, visible spatial/state correctness, re-observation support, and re-observed consistency. The complete annotation inventory covers 2,641 annotation rows across five semantic families; the released subset contains 2,547 deduplicated annotator verdicts over 1,156 comparison pairs. Table~\ref{tab:human-release-registry} records the release registry by annotation family, and Table~\ref{tab:human-agreement-metrics} reports inter-annotator agreement by diagnostic dimension.

\begin{table}[t]
\centering
\caption{\textbf{Human annotation release registry.}
Rows list annotation families, release roles, row counts, label counts, and uses.}
\label{tab:human-release-registry}
\WRTableSetup
\begin{tabularx}{\linewidth}{@{}L{0.11\linewidth} L{0.20\linewidth} L{0.16\linewidth} L{0.16\linewidth} X@{}}
\toprule
\WRTableHead
Family & Role & Rows & Labels & Use \\
\midrule
Main & Reliability anchor & 360 (355 active) & 4-way & Human-human agreement. \\
Supp. & Robustness & 290 & 3-way & Agreement checks. \\
Src/geo & Model slices & 276 & 138 triples & Source/geometry diagnostics. \\
Broad & Case mining & 1418 & 401 triples & Hard-pair evidence. \\
Stratified & Re-observed coverage & 297 & 297 triples & Re-observed-metric probes. \\
\bottomrule
\end{tabularx}
\end{table}

\begin{table}[t]
\centering
\caption{\textbf{Human agreement by diagnostic dimension.}
Rows report $\kappa$, $\alpha$, and AC1 for the main and supplementary annotation sets.}
\label{tab:human-agreement-metrics}
\WRTableTightSetup{3pt}
\begin{tabular}{@{}lcccccc@{}}
\toprule
\WRTableHead
Metric & \multicolumn{3}{c}{Main set} & \multicolumn{3}{c}{Supp. set} \\
\cmidrule(lr){2-4}\cmidrule(lr){5-7}
& $\kappa$ & $\alpha$ & AC1 & $\kappa$ & $\alpha$ & AC1 \\
\midrule
Requested view & 0.535 & 0.485 & 0.898 & 0.524 & 0.568 & 0.677 \\
Visual integrity & 0.536 & 0.468 & 0.877 & 0.546 & 0.644 & 0.700 \\
Visible spatial & 0.316 & 0.214 & 0.788 & 0.188 & 0.307 & 0.481 \\
Visible event-state & 0.504 & 0.424 & 0.790 & 0.497 & 0.614 & 0.608 \\
Re-observed spatial & 0.632 & 0.560 & 0.875 & 0.403 & 0.716 & 0.880 \\
Re-observed event-state & 0.620 & 0.433 & 0.937 & 0.409 & 0.754 & 0.880 \\
\bottomrule
\end{tabular}
\end{table}

\begin{table}[t]
\centering
\caption{\textbf{World-state evaluator--human bridge.}
Decision $\kappa$ is thresholded agreement, Rank $\rho$ and Holdout $\rho$ are Spearman alignment, and J, NJ, and Bal. denote judgeable, non-judgeable, and balanced judgeability agreement for re-observed metrics.}
\label{tab:evaluator-human-validation}
\WRTableTightSetup{3pt}
\begin{tabularx}{\linewidth}{@{}L{0.27\linewidth} c c c c c c@{}}
\toprule
\WRTableHead
Metric & \multicolumn{3}{c}{Evaluator agreement} & \multicolumn{3}{c}{Judgeability agree.} \\
\cmidrule(lr){2-4}\cmidrule(lr){5-7}
& Decision $\kappa$ & Rank $\rho$ & Holdout $\rho$ & J & NJ & Bal. \\
\midrule
Visible spatial & 0.199 & 0.386 & 0.360 & -- & -- & -- \\
Visible event-state & 0.271 & 0.484 & 0.444 & -- & -- & -- \\
Re-observed spatial & 0.388 & 0.667 & 0.844 & 0.941 & 0.966 & 0.954 \\
Re-observed event-state & 0.427 & 0.660 & 0.775 & 0.941 & 0.946 & 0.943 \\
\bottomrule
\end{tabularx}
\end{table}

Table~\ref{tab:evaluator-human-validation} reports the world-state evaluator--human bridge that underlies these checks. The fixed DINOv2 visual-integrity proxy is calibrated separately from the world-state VLM probes, with exact/strict/tie pairwise agreement of 0.730/0.773/0.702, weighted $\kappa=0.678$, and rank $\rho=0.709$ on a 190-pair family holdout. The world-state bridge uses 230 pairwise comparisons over 261 unique videos; the recovered 2026-06-05 measurement-available bridge has effective visible-spatial / visible-event-state / re-observed-spatial / re-observed-event-state denominators of 228/229/136/136. These bridge denominators are distinct from the 645-pair main human-comparison reliability set and the 2,547-verdict released human-data subset.

The thresholded decision check separates ranking from calibration. Under the reported evaluator configuration, visible spatial and visible event-state metrics introduce no human-preference-to-opposite reversals, while re-observed spatial and re-observed event-state contain one and eight reversals, respectively. Re-observed-metric scores remain conditional on judgeability: agreement on whether hidden-and-returned evidence exists is reported separately from conditional re-observed-consistency accuracy.

\section{Preference-Pair Export and Reward/Policy Outlook}
\label{app:preference-pair-export}

The same decoupled metric records can be exported as preference pairs. This appendix specifies how WRBench records can be converted into preference pairs for future reward-model or policy-training experiments. It documents the export interface only; the main paper does not use preference optimization, downstream reward-model training, DPO, or policy-improvement results as evidence. Table~\ref{tab:preference-pair-export-summary} summarizes the export counts after missing assets are removed.


\begin{table}[t]
\centering
\caption{\textbf{Preference-pair export summary.} Export counts after removing missing assets.}
\label{tab:preference-pair-export-summary}
\WRTableSetup
\begin{tabular}{l r r r}
\toprule
\WRTableHead
Set & Input & Missing & Kept \\
\midrule
Stage 1 & 880 & 39 & 841 \\
Ext. 1 & 900 & 135 & 765 \\
Ext. 2 & 900 & 0 & 900 \\
\midrule
Total & 2680 & 174 & 2506 \\
\bottomrule
\end{tabular}
\end{table}

\subsection{Reward and Policy-Training Outlook}
\label{app:reward-policy-outlook}

The export interface turns WRBench diagnostics into reward and policy-training inputs. A reward model can mine pairs from the same diagnostic dimensions, for example preferring stronger target-relative camera displacement when visual integrity and returned-evidence judgeability are preserved. A policy-training study can then test whether optimizing those pairs changes held-out WRBench evidence dimensions.

The present paper provides the measurement handoff: metric records, judgeability flags, pair construction rules, and dynamic-versus-dynamic filtering. Reward optimization and policy training become downstream tests over the same evidence graph. The diagnostic structure matters because rewarding visible plausibility alone could reinforce exactly the failure WRBench exposes: keeping the salient subject in frame while failing to create a judgeable hidden-and-returned world-state test.

\subsection{Dense-Control and Future Benchmark Extensions}
\label{app:dense-control-future}

Dense-control world-model interfaces should be added to WRBench only with leakage-aware settings. If target-view depth, segmentation, edge maps, LiDAR, HD maps, or other dense controls already encode the post-event endpoint, a high score can reflect control following rather than hidden-state inference. This is why Cosmos-style or other dense-control systems should enter as future model clusters with explicit input-policy labels.

Three settings separate the claims:
\begin{enumerate}
  \item \textbf{Source-only control.} Controls are extracted from the source view or source video only, so the target-view endpoint is not supplied.
  \item \textbf{Endpoint-masked target control.} Target-view controls are allowed, but the object/contact endpoint region is masked or ambiguous.
  \item \textbf{Full target control.} Target-view dense controls provide an upper-bound control-following condition, not evidence of internal world-state persistence.
\end{enumerate}

Reward work or policy training should follow the same separation. WRBench records can mine preference pairs over target-relative camera displacement, visual integrity, judgeable re-observation, and re-observed consistency, but model-gain claims require held-out evidence after training.

\end{document}